\documentclass[11pt]{article}

\usepackage[preprint]{acl}

\usepackage{times}
\usepackage{latexsym}
\usepackage[T1]{fontenc}
\usepackage[utf8]{inputenc}
\usepackage{microtype}
\usepackage{inconsolata}
\usepackage{graphicx}
\usepackage{booktabs}
\usepackage{amsmath}
\usepackage{amssymb}
\usepackage{array}
\usepackage{tabularx}
\usepackage{colortbl}
\usepackage{placeins}
\usepackage{pifont}
\usepackage{tikz}
\usepackage{xspace}
\usepackage{upgreek}
\newsavebox{\statebox}
\makeatletter
\newcommand\appendixautoref{%
  \let\Hy@app@orgrsc\hyper@refstepcounter
  \def\hyper@refstepcounter##1{\edef\Hy@app@tmp{##1}%
    \ifnum0\ifx\Hy@app@tmp\Hy@app@sec1\fi\ifx\Hy@app@tmp\Hy@app@sub1\fi>0
      \MakeLinkTarget[appendix]{##1}\else\Hy@app@orgrsc{##1}\fi}}
\def\Hy@app@sec{section}\def\Hy@app@sub{subsection}
\makeatother
\definecolor{oursbg}{gray}{0.92}
\definecolor{refbg}{gray}{0.95}
\definecolor{grpbg}{HTML}{E9EFF6}
\newcommand{\grp}[2]{\rowcolor{grpbg}\multicolumn{#1}{c}{\textit{#2}}}
\newcommand{\dlt}[1]{\,{\scriptsize\textcolor{black!55}{#1}}}
\definecolor{dltpos}{HTML}{1E8449}% muted green for gains
\definecolor{dltneg}{HTML}{C0392B}% muted red for losses
\newcommand{\dltp}[1]{\,{\scriptsize\textcolor{dltpos}{$\uparrow$#1}}}
\newcommand{\dltn}[1]{\,{\scriptsize\textcolor{dltneg}{$\downarrow$#1}}}

\newcommand{\trSS}{\ensuremath{\mathrm{S}{\to}\mathrm{S}}}
\newcommand{\trSW}{\ensuremath{\mathrm{S}{\to}\mathrm{W}}}

\newcommand{\trWW}{\ensuremath{\mathrm{W}{\to}\mathrm{W}}}
\newcommand{\taur}{\texttt{$\uptau^2$-retail}}
\newcommand{\taub}{\texttt{$\uptau^2$-bench}}
\newcommand{\MethodName}{\textsc{StaGPO}\xspace}
\newcommand{\sig}[1]{\textbf{#1}}
\definecolor{rqone}{HTML}{C2410C}\definecolor{rqtwo}{HTML}{1F6FD1}\definecolor{rqthree}{HTML}{0E8C6B}
\DeclareRobustCommand{\rqtag}[2]{\tikz[baseline=(rq.base)]\node[rounded corners=2pt,fill=#2!10,draw=#2!75,line width=0.5pt,inner xsep=2.2pt,inner ysep=1.3pt,text=#2!85!black,font=\sffamily\bfseries\footnotesize](rq){RQ#1};}% research-question badge
\newcommand{\sd}[1]{{\scriptsize$\,\pm$#1}}

\title{Not Every User Says It All: Training Tool-Using Agents Beyond Ideal Simulated Users}

\author{Han Bao\textsuperscript{1} \quad Zheyuan Zhang\textsuperscript{1} \quad Kaiwen Shi\textsuperscript{1} \quad Yanfang Ye\textsuperscript{1} \\
  \textsuperscript{1}University of Notre Dame \\
  \texttt{\{hbao, yye7\}@nd.edu}}


\begin{document}
\maketitle

\begin{abstract}
Tool-using agents are trained and evaluated almost exclusively with \emph{strong} simulated users who state every detail clearly, yet real users are often weaker. We construct weak users by masking sentences of a simulated user's replies, holding the task and the user's knowledge fixed, and cross-evaluate agents trained with strong and weak users on \taub{} and \texttt{ToolWOZ}. Weak users lower the success of every policy, and training with them does not help: no policy trained with weak users, through fixed masks, mixtures or a curriculum, serves weak users better than the strong-trained policy beyond run-to-run noise, and prompting the agent to ask more does not help either. We propose \MethodName (\underline{Sta}te-\underline{G}rouped \underline{P}olicy \underline{O}ptimization): the agent writes what the task needs, what is missing and how it is being filled, and \textsc{GRPO} uses this state to compare turns of different rollouts that reach the same situation, assigning step-level credit. Within the standard recipe of a small SFT cold start followed by RL, \MethodName makes weak-user training useful: on \texttt{ToolWOZ}, the \texttt{Qwen3.5-2B} policy trained with weak users reaches 76.8\% with weak users, the highest of any \texttt{Qwen3.5-2B} policy we trained, 4.8 points above \textsc{GRPO} and 2.8 above the strongest baseline; it is also 5.4 points above \textsc{GRPO} with strong users, and generalises to an unseen mask rate; trained with strong users instead, it serves weak users at least as well as \textsc{GRPO} trained with weak users. On \taub{} it scores highest with weak users for both models, by a small margin for \texttt{Qwen3.5-4B}.
\end{abstract}

\section{Introduction}

Interaction with users is now central to tool-using language agents. Benchmarks pair the agent with an LLM-simulated customer who holds a task \citep{yao2024taubench,barres2025tau2,lattimer2024sparse}, and reinforcement learning trains agents against such simulators \citep{qian2025userrl,zhou2025sweetrl}. The simulated user is almost always cooperative and fully informative, a \emph{strong} user, both in training and at evaluation, although alignment to an idealised average user is known to break at the edges of the user distribution \citep{bao2026edge}. Real users are often weaker: they leave out a constraint, answer half of a two-part question, or speak through a channel that drops part of what they said, and agents tend to act on what they have rather than notice what they lack \citep{bao2026driftbench,shi2026confidence, shi2026sage}. Recent work evaluates agents against harder simulated users \citep{shim2025noncollab,laban2025lost}, but how agents should be \emph{trained} when their users are weak has not been studied. We study this by withholding parts of a cooperative simulator's replies (\autoref{sec:setup}) and training Qwen3.5 policies \citep{qwen2026qwen35,yang2025qwen3} with \textsc{GRPO} \citep{shao2024deepseekmath} on \taub{} \citep{barres2025tau2} and \texttt{ToolWOZ} \citep{lattimer2024sparse} under every combination of strong and weak users.

\begin{figure*}[t]
\centering
\includegraphics[width=\textwidth]{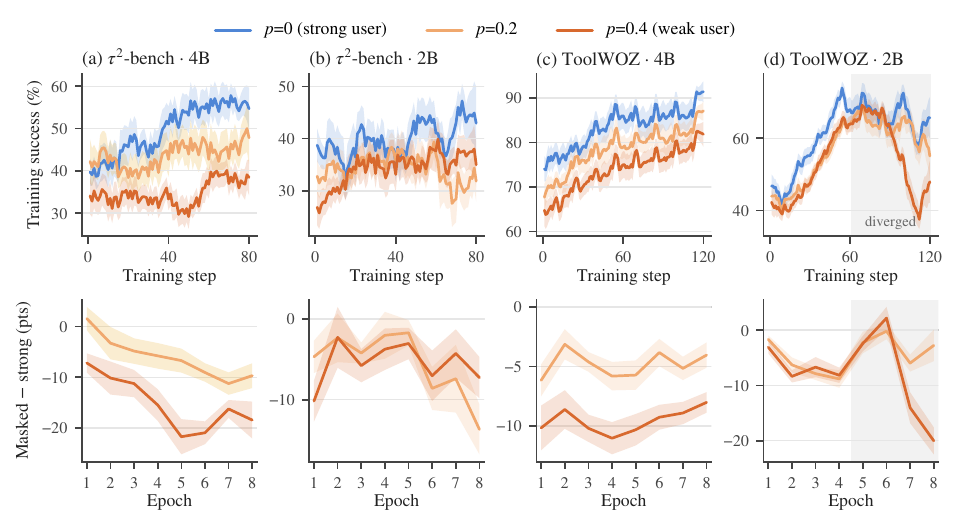}
\caption{\textbf{Training with strong and weak users} on \taub{} and \texttt{ToolWOZ}, \texttt{Qwen3.5-4B} and \texttt{2B}. Top: training success of each arm (moving average); bottom: success of each masked arm minus the unmasked arm, per epoch (shading $\pm$1\,SE; grey: the \texttt{ToolWOZ} 2B arms diverge).}
\label{fig:training}
\end{figure*}

\paragraph{Training with weak users does not help.} Weak users lower the success of every policy; on \taub{} the 4B policy trained with strong users loses 12.7 points when its user is masked at $p{=}0.4$ (\autoref{sec:gap}, \autoref{fig:training}). The standard remedy, training under the mismatch, does not help: no policy trained with weak users, through fixed masks, mixtures or a curriculum, serves weak users better than the strong-trained policy beyond run-to-run noise, on either benchmark or model size, and several lose strong-user performance. With \textsc{GRPO}, we find no gain from weak-user dialogues beyond noise.

\paragraph{From asking to grouping.} A prompt that makes the agent ask five times as often after a masked turn leaves success unchanged, and in counterfactual forks a single request at the right point recovers far more than asking whenever something is masked (\autoref{sec:gap}). The agent does not recognise what is missing: it flags about half of the details it has received as missing, and being told which fields are missing does not make it act on them. Rewarding repairs, or rewarding a written state for its accuracy, did not help in our earlier experiments either: repair rewards were too sparse to change behaviour, and a state score, unrelated to success within a group, lowered the training reward when added to it (\autoref{app:path}). What a written state can do is tell which turns of different rollouts are in the same situation.

\paragraph{Teaching need, gap and fill.} We therefore use the state as a grouping key rather than a training target and propose \emph{state-grouped policy optimization} (\MethodName; \autoref{sec:method}). Before every action the agent writes down what the task needs from the user, what is still missing and how it is being filled. \textsc{GRPO} gives every turn of a dialogue the same advantage; \MethodName compares turns of different rollouts that reach the same need--gap--fill situation by their discounted return, a step-level signal without a critic. Following \textsc{UserRL} \citep{qian2025userrl}, we train from a small SFT cold start. Evaluated with the user it is trained with, \MethodName is above or level with \textsc{GRPO} at every training mask on \texttt{ToolWOZ}, for \texttt{Qwen3.5-2B} and 4B, and 4.8 points above it with weak users (2B; \autoref{sec:rq4}). Its gains transfer across users: trained with weak users it is the best policy with strong users and at an unseen mask rate, and trained with strong users it serves weak users at least as well as \textsc{GRPO} trained with weak users (\autoref{sec:rq5}). On \taub{} it scores highest with weak users for both models, by a small margin for \texttt{Qwen3.5-4B}.

In summary, we contribute (i) a controlled weak-user protocol that degrades only the channel and records what was said and what was heard; (ii) a train-by-test cross-evaluation on two benchmarks and two model sizes showing that, with \textsc{GRPO}, weak-user training does not serve weak users better than strong-user training, and that asking more is not the remedy; and (iii) \MethodName, which assigns step-level credit through a need--gap--fill state the agent writes and, from a small SFT cold start, makes weak-user training useful on \texttt{ToolWOZ}, with gains that transfer across users.

\begin{figure*}[t]
\centering
\includegraphics[width=\textwidth]{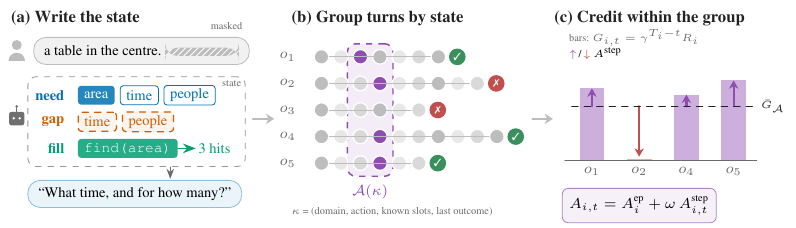}
\caption{\textbf{\MethodName.} (a) The agent writes a hidden state: what the task \textcolor[HTML]{0072B2}{needs}, what is still missing (\textcolor[HTML]{D55E00}{gap}) and how it is being \textcolor[HTML]{009E73}{filled}. (b) Turns of different rollouts with the same state key $\kappa$ form an anchor group. (c) Each anchored turn is credited by its return relative to the group mean, added to \textsc{GRPO}'s episode advantage.}
\label{fig:overview}
\end{figure*}

\section{Related Work}

\paragraph{Multi-turn agent RL with simulated users.} Reinforcement learning for multi-turn language agents ranges from hierarchical value methods \citep{zhou2024archer} to outcome-reward policy optimisation \citep{wang2025ragen}, including agents trained against LLM-simulated users \citep{zhou2025sweetrl,qian2025userrl}. Step-level credit has come from critics with privileged information \citep{pinto2017asymmetric,zhou2025sweetrl} or from grouping steps that reach an identical environment state \citep{feng2025gigpo}, which dialogues with a free-text user rarely do; \MethodName instead groups turns by a state the agent writes itself. We hold the algorithm, reward and task fixed and vary only how much of the user's reply reaches the agent.

\paragraph{Harder simulated users.} Tool-agent benchmarks pair the agent with a simulated user that holds the full task \citep{yao2024taubench,barres2025tau2,lattimer2024sparse,wang2023mint,lu2024toolsandbox}. Non-collaborative simulators \citep{shim2025noncollab} and instructions sharded across turns \citep{laban2025lost} make the user harder through prompting and evaluate fixed models; we degrade the channel after the user speaks, keep the task fixed, and ask whether \emph{training} with the weaker user helps.

\paragraph{Underspecification and clarification.} Agents struggle with vague or underspecified requests \citep{qian2024tellmemore,wang2024learningtoask,vijayvargiya2025ambigswe,bao2026driftbench}, and training methods reward useful questions \citep{andukuri2024stargate,chen2024learning,wu2025collabllm,kong2026infopo}. There the missing information is absent from the request; in our setting the user said it but it did not arrive, and asking more recovers little.

\paragraph{Training under perturbation.} Robustness to perturbed inputs is a core axis of trustworthy model evaluation \citep{huang2026trustgen}, and training under a perturbation is the standard remedy for a train--test mismatch \citep{tobin2017domain,liu2020laug,chen2026noisyagent}. Our perturbation removes information the reward depends on, and fixed, mixed and curriculum schedules do not serve weak users better than strong-user training. An extended discussion is in \autoref{app:related}.

\section{Problem Setup}
\label{sec:setup}

\begin{table*}[t]
\centering
\setlength{\tabcolsep}{2pt}
\renewcommand{\dltp}[2][12]{\cellcolor{dltpos!#1}{\footnotesize\textcolor{dltpos!80!black}{$\uparrow$#2}}}%
\renewcommand{\dltn}[2][12]{\cellcolor{dltneg!#1}{\footnotesize\textcolor{dltneg!80!black}{$\downarrow$#2}}}%
\begin{tabularx}{\textwidth}{l*{6}{>{\raggedleft\arraybackslash}X@{\hspace{4pt}}>{\raggedright\arraybackslash}p{0.95cm}}}
\toprule
 & \multicolumn{6}{c}{\taub{}} & \multicolumn{6}{c}{\texttt{ToolWOZ}} \\
\cmidrule(lr){2-7}\cmidrule(lr){8-13}
Trained with & \multicolumn{2}{c}{$p{=}0$} & \multicolumn{2}{c}{0.2} & \multicolumn{2}{c}{0.4} & \multicolumn{2}{c}{$p{=}0$} & \multicolumn{2}{c}{0.2} & \multicolumn{2}{c}{0.4} \\
\midrule
\grp{13}{Qwen3.5-4B} \\
none (untrained) & 46.7 &  & 40.5 & \dltn[22]{6.2} & 32.0 & \dltn[42]{14.7} & 77.3 &  & 72.6 & \dltn[18]{4.7} & 69.0 & \dltn[27]{8.3} \\
\rowcolor{refbg} strong user & 53.9 &  & 51.0 & \dltn[14]{2.9} & 41.2 & \dltn[37]{12.7} & 85.5 &  & 80.5 & \dltn[19]{5.0} & 74.6 & \dltn[33]{10.9} \\
weak, $p{=}0.2$ & 45.1 &  & 39.2 & \dltn[21]{5.9} & 40.2 & \dltn[19]{4.9} & 84.0 &  & 82.4 & \dltn[11]{1.6} & 77.1 & \dltn[24]{6.9} \\
weak, $p{=}0.4$ & 47.1 &  & 42.2 & \dltn[19]{4.9} & 39.2 & \dltn[26]{7.8} & 85.9 &  & 80.5 & \dltn[20]{5.4} & 74.8 & \dltn[34]{11.1} \\
\midrule
\grp{13}{Qwen3.5-2B} \\
none (untrained) & 41.2 &  & 33.0 & \dltn[27]{8.2} & 30.7 & \dltn[32]{10.5} & 49.0 &  & 42.6 & \dltn[22]{6.4} & 40.7 & \dltn[27]{8.3} \\
\rowcolor{refbg} strong user & 40.2 &  & 36.6 & \dltn[16]{3.6} & 33.3 & \dltn[24]{6.9} & 74.9 &  & 69.2 & \dltn[21]{5.7} & 63.7 & \dltn[34]{11.2} \\
weak, $p{=}0.2$ & 34.3 &  & 29.4 & \dltn[19]{4.9} & 26.5 & \dltn[26]{7.8} & 76.1 &  & 74.8 & \dltn[10]{1.3} & 66.8 & \dltn[29]{9.3} \\
weak, $p{=}0.4$ & 44.4 &  & 40.5 & \dltn[16]{3.9} & 38.9 & \dltn[20]{5.6} & 71.2 &  & 67.7 & \dltn[15]{3.5} & 62.1 & \dltn[29]{9.1} \\
\bottomrule
\end{tabularx}
\caption{\textbf{Cross-evaluation of policies trained with strong and weak users.} Success (\%) by the user's mask at evaluation ($p{=}0$: strong user); rows: the user at training; small numbers to the right: change from $p{=}0$. Standard deviations: \autoref{tab:main_full}.}
\label{tab:main}
\end{table*}

\paragraph{Weak users by channel masking.} Prompting a simulator to be vaguer also changes what it knows, what it wants and how it behaves, so any change in success would mix several factors; swapping it for a smaller or larger model does the same. In an exploratory study (\autoref{app:scale}), policies trained with \texttt{gpt-oss-20b} or \texttt{gpt-oss-120b} were first better with their own simulator (by 3.9 and 8.8 points after four epochs), but the ordering had reversed after eight, so simulator size gives no fixed ranking of users. We instead weaken the user \emph{after} it speaks, holding the task, the user's knowledge and goals, the reward and the training recipe fixed. Each user reply is split into sentences; each sentence is replaced by \texttt{[inaudible]} independently with probability $p\in\{0,0.2,0.4\}$ (auxiliary runs use 0.15 and 0.30). Control markers that end a conversation are never masked but count toward the denominator, so the realised rate is slightly below $p$ (e.g.\ 0.37--0.39 at $p{=}0.4$). The user simulator is unchanged and still sees its own unmasked history, so, like a real user who believes they were clear, it does not know what the agent missed; only the agent receives the degraded text. Because we know both what was said and what was heard, we can trace each task value through the channel and fork a dialogue where information was lost (\autoref{app:ask}). We write \textsc{S} for the unmasked (strong) user and \textsc{W} for a masked (weak) one, and \textsc{A}$\to$\textsc{B} for a policy trained with \textsc{A} and evaluated with \textsc{B}.

\section{State-Grouped Policy Optimization}
\label{sec:method}

With a weak user, success depends on whether the agent tracks what the task needs, what is still missing and how it is being filled. State-grouped policy optimization (\MethodName; \autoref{fig:overview}) has the agent write this down and uses it to assign credit to individual turns.

\paragraph{Multi-turn \textsc{GRPO}.} For a task $q$, the policy $\pi_\theta$ samples $G$ rollouts; at turn $t$ of rollout $i$ it generates $y_{i,t}\sim\pi_\theta(\cdot\mid h_{i,t})$ given the dialogue history, and the rollout ends with reward $R_i\in[0,1]$. \textsc{GRPO} \citep{shao2024deepseekmath} normalises rewards within the group, $A^{\text{ep}}_i=(R_i-\mu_q)/(\sigma_q+\epsilon)$, and maximises
\begin{equation}
\mathcal{J}(\theta)=\mathbb{E}\Big[\frac{1}{N}\sum_{\ell}\min\big(\rho_\ell A_\ell,\,\tilde\rho_\ell A_\ell\big)\Big]-\beta\,\mathbb{D}_{\text{KL}},
\label{eq:obj}
\end{equation}
where $\ell$ runs over the $N$ generated tokens, $\rho_\ell$ is the token's importance ratio and $\tilde\rho_\ell$ its clipped value \citep{yu2025dapo}. A token of turn $y_{i,t}$ receives $A_\ell=A_{i,t}$, and \textsc{GRPO} sets $A_{i,t}=A^{\text{ep}}_i$ for all $t$: a turn that asks for the missing date and a turn that repeats a failed search share one credit.

\paragraph{Need--gap--fill state.} Each turn is written as $y_{i,t}=(z_{i,t},a_{i,t})$: a JSON state $z$ in \texttt{<state>} tags, hidden from the user, followed by an action $a$. The state $z=(n,g,f)$ records the \emph{need} $n$ (goal, pending task, requirements, entities), the \emph{gap} $g$ (user information still missing) and the \emph{fill} $f$ (the last three tool calls and their outcomes); \autoref{fig:state} shows an example.

\begin{figure}[t]
\centering
\begin{lrbox}{\statebox}
\begin{minipage}{0.94\columnwidth}
\scriptsize
\begin{verbatim}
<state>{"goal": "book a Thai restaurant for 4",
 "pending_task": "restaurant.book",
 "requirements": {"area": "centre", "food": "thai",
                  "people": 4, "day": null, "time": null},
 "missing_user_information": ["day", "time"],
 "attempts": ["search(area, food) -> 3 results"]}</state>
\end{verbatim}
\vspace{-6pt}
\small $\kappa=$ (restaurant, book, \{area, food, people\}, results)
\end{minipage}
\end{lrbox}
\setlength{\fboxsep}{4pt}\fcolorbox{gray!35}{gray!6}{\usebox{\statebox}}
\caption{\textbf{A written state} (\texttt{ToolWOZ}, abridged, constructed) and its anchor key (\autoref{eq:key}). Slot values are not part of the key, so rollouts that know different areas or party sizes share it.}
\label{fig:state}
\end{figure}

\paragraph{Anchor groups.} With free-text users two rollouts almost never reach an identical history, so grouping by identical environment states \citep{feng2025gigpo} rarely applies. Instead, a fixed abstraction of the written state defines when two turns are in the same situation,
\begin{equation}
\kappa_{i,t}=\phi(z_{i,t})=\big(d,\,c,\,\operatorname{slots}(n),\,o(f)\big),
\label{eq:key}
\end{equation}
the pending domain and action, the requirement slots already known, and the outcome of the last tool call; the known slots determine the gap, since the missing slots are the remaining requirements. Turns of one task with equal keys form an anchor group $\mathcal{A}_q(\kappa)=\{(i,t):\kappa_{i,t}=\kappa\}$, kept if they come from at least $m$ rollouts.

\paragraph{Step advantage.} With the discounted return $G_{i,t}=\gamma^{\,T_i-t}R_i$, a turn in an anchor group receives
\begin{gather}
A^{\text{step}}_{i,t}=\operatorname{clip}\!\Big(\frac{G_{i,t}-\bar G_{\mathcal{A}}}{\sigma_q+\epsilon},\,\pm c\Big),
\label{eq:step}\\
A_{i,t}=A^{\text{ep}}_i+\omega\,A^{\text{step}}_{i,t},
\label{eq:adv}
\end{gather}
and other turns keep $A^{\text{ep}}_i$; \autoref{eq:obj} is otherwise unchanged. Dividing by the task's reward spread $\sigma_q$ rather than the anchor group's keeps both terms on one scale (a two-member group would always give $\pm1$), and the discount prefers turns that reach the outcome sooner from the same situation. We use $\gamma{=}0.95$, $m{=}3$, $c{=}3$, $\omega{=}1$; with the cold start, the share of anchored turns grows from 30\% early in training to 48\% at the end (\autoref{tab:cold}).

\paragraph{Interpretation.} $\bar G_{\mathcal{A}}$ estimates the value of the situation $\kappa$, so $A^{\text{step}}_{i,t}\propto\hat Q(\kappa_{i,t},y_{i,t})-\hat V(\kappa_{i,t})$: unlike \textsc{GRPO}'s task-level baseline, it removes the part of the return explained by where a rollout ended up (e.g.\ whether the user's date was lost) and keeps the part explained by what the agent did there.

\section{Experimental Setup}
\label{sec:expsetup}

\paragraph{Environments.} We use the retail domain of \taub{} \citep{barres2025tau2}, in which the user has no tools and success reflects the agent's actions, and \texttt{ToolWOZ} \citep{lattimer2024sparse}, booking tasks over restaurant, hotel, train and attraction APIs. In all main experiments the user is \texttt{gpt-oss-120b} \citep{openai2025gptoss} and evaluation measures binary task success. Task splits are in \autoref{app:env}.

\paragraph{Training.} We train \texttt{Qwen3.5-4B} and \texttt{Qwen3.5-2B} in non-thinking mode with LoRA \citep{hu2022lora} in verl \citep{sheng2024hybridflow}, using a learning rate of $3{\times}10^{-5}$ for both \textsc{GRPO} and \MethodName across both benchmarks and model sizes. The cross-evaluation (RQ1) trains \textsc{GRPO} from the base model with fixed masks and with two mixed schedules: \emph{hybrid}, which rotates masks 0, 0.2 and 0.4 over training samples, and \emph{curriculum}, whose mask grows from 0 to 0.4. As in \textsc{UserRL} \citep{qian2025userrl}, \MethodName starts from a small SFT cold start on 33 of the model's own training dialogues, with no stronger model involved. We compare it with \textsc{GRPO}, with \textsc{UserRL}, which starts from the same cold start and replaces only the advantage, and with \textsc{InfoPO} \citep{kong2026infopo}, trained as in its paper. Training rewards, hyperparameters and baseline details are specified in \autoref{app:hparams}.

\paragraph{Evaluation.} The simulated user cannot be seeded, so repeated evaluations of a policy differ; we compare policies by \emph{task-paired} differences (\autoref{app:stats}); in \autoref{tab:sg} and \autoref{tab:sgmask}, bold marks the best value per column. Where checkpoints are selected, the selection uses validation tasks under the training mask among steps fixed in advance. Evaluation sets, seeds and selection rules are in \autoref{app:stats}.

\section{Experiments}
\label{sec:exp}

Our experiments follow the questions a practitioner faces when users may be weak. Training against simulated users implicitly assumes that the simulator's strength does not matter, and the standard remedy for a train--test mismatch is to train under it; we therefore first ask whether standard RL training, with strong users or with weak ones, serves weak users, and what the agent lacks when it does not \rqtag{1}{rqone}. The answer points to tracking what the task needs, what is missing and how it is being filled, which \MethodName makes explicit. We then ask whether \MethodName helps when the policy is trained and evaluated with the same user \rqtag{2}{rqtwo}, and whether its gains carry over when the user at evaluation differs from the user at training, as it will after deployment \rqtag{3}{rqthree}. RQ1 uses the cross-evaluation of \autoref{tab:main}, RQ2--3 the \MethodName study of \autoref{tab:sg} and \autoref{tab:sgmask}.

\subsection{\texorpdfstring{\rqtag{1}{rqone}\ }{RQ1: }Can standard RL training serve weak users?}
\label{sec:gap}
\label{sec:training}

\begin{figure}[t]
\centering
\includegraphics[width=\columnwidth]{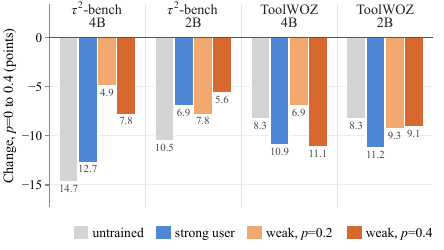}
\caption{\textbf{Change in success when the user becomes weak.} Success at $p{=}0.4$ minus success at $p{=}0$ (points) for every policy of \autoref{tab:main}. On \taub{} 4B the weak-trained policies lose less because they start lower with the strong user, not because they do better with the weak one.}
\label{fig:maskcost}
\end{figure}

\begin{table*}[t]
\centering
\setlength{\tabcolsep}{2pt}
\renewcommand{\dltp}[2][12]{\cellcolor{dltpos!#1}{\footnotesize\textcolor{dltpos!80!black}{$\uparrow$#2}}}%
\renewcommand{\dltn}[2][12]{\cellcolor{dltneg!#1}{\footnotesize\textcolor{dltneg!80!black}{$\downarrow$#2}}}%
\begin{tabularx}{\textwidth}{l*{6}{>{\raggedleft\arraybackslash}X@{\hspace{4pt}}>{\raggedright\arraybackslash}p{0.95cm}}}
\toprule
 & \multicolumn{6}{c}{\taub{}} & \multicolumn{6}{c}{\texttt{ToolWOZ}} \\
\cmidrule(lr){2-7}\cmidrule(lr){8-13}
Evaluated at & \multicolumn{2}{c}{$p{=}0$} & \multicolumn{2}{c}{0.2} & \multicolumn{2}{c}{0.4} & \multicolumn{2}{c}{$p{=}0$} & \multicolumn{2}{c}{0.2} & \multicolumn{2}{c}{0.4} \\
\midrule
\grp{13}{Qwen3.5-2B} \\
Untrained & 41.2 &  & 33.0 &  & 30.7 &  & 49.0 &  & 42.6 &  & 40.7 &  \\
\midrule
\textsc{GRPO} & 34.4 &  & 30.2 &  & 24.0 &  & 78.7 &  & 74.3 &  & 72.0 &  \\
\textsc{Hybrid} & 30.7 & \dltn[16]{3.7} & 29.7 & \dltn[8]{0.5} & 27.6 & \dltp[16]{3.6} & 64.0 & \dltn[42]{14.7} & 57.1 & \dltn[42]{17.2} & 52.0 & \dltn[42]{20.0} \\
\textsc{Curriculum} & \textbf{45.8} & \dltp[34]{11.4} & 33.3 & \dltp[14]{3.1} & 33.3 & \dltp[29]{9.3} & 65.7 & \dltn[38]{13.0} & 65.1 & \dltn[29]{9.2} & 62.9 & \dltn[29]{9.1} \\
\textsc{UserRL} & 31.2 & \dltn[15]{3.2} & 27.1 & \dltn[14]{3.1} & 28.1 & \dltp[17]{4.1} & 61.4 & \dltn[42]{17.3} & 60.4 & \dltn[40]{13.9} & 57.9 & \dltn[41]{14.1} \\
\textsc{InfoPO} & 35.4 & \dltp[9]{1.0} & \textbf{37.5} & \dltp[25]{7.3} & 30.2 & \dltp[22]{6.2} & 78.3 & \dltn[8]{0.4} & 75.7 & \dltp[10]{1.4} & 74.0 & \dltp[12]{2.0} \\
\rowcolor{oursbg} \MethodName & \textbf{45.8} & \dltp[34]{11.4} & 36.5 & \dltp[22]{6.3} & \textbf{40.6} & \dltp[42]{16.6} & \textbf{84.1} & \dltp[20]{5.4} & \textbf{81.5} & \dltp[24]{7.2} & \textbf{76.8} & \dltp[19]{4.8} \\
\midrule
\grp{13}{Qwen3.5-4B} \\
Untrained & 46.7 &  & 40.5 &  & 32.0 &  & 77.3 &  & 72.6 &  & 69.0 &  \\
\midrule
\textsc{GRPO} & 47.1 &  & 42.2 &  & 39.2 &  & 85.9 &  & 80.5 &  & 74.8 &  \\
\textsc{Hybrid} & 40.6 & \dltn[23]{6.5} & \textbf{47.9} & \dltp[21]{5.7} & 41.7 & \dltp[13]{2.5} & 86.1 & \dltp[7]{0.2} & 82.7 & \dltp[12]{2.2} & 77.8 & \dltp[14]{3.0} \\
\textsc{Curriculum} & 44.8 & \dltn[13]{2.3} & 44.8 & \dltp[13]{2.6} & 36.5 & \dltn[13]{2.7} & 86.5 & \dltp[8]{0.6} & 80.1 & \dltn[8]{0.4} & 75.4 & \dltp[8]{0.6} \\
\textsc{UserRL} & 45.2 & \dltn[12]{1.9} & 43.6 & \dltp[10]{1.4} & 40.1 & \dltp[9]{0.9} & 69.2 & \dltn[42]{16.7} & 66.0 & \dltn[42]{14.5} & 59.1 & \dltn[42]{15.7} \\
\textsc{InfoPO} & 48.0 & \dltp[9]{0.9} & 44.5 & \dltp[13]{2.3} & 40.7 & \dltp[11]{1.5} & 83.8 & \dltn[12]{2.1} & 84.6 & \dltp[17]{4.1} & 77.9 & \dltp[14]{3.1} \\
\rowcolor{oursbg} \MethodName & \textbf{51.0} & \dltp[16]{3.9} & 44.8 & \dltp[13]{2.6} & \textbf{42.7} & \dltp[15]{3.5} & \textbf{87.1} & \dltp[10]{1.2} & \textbf{85.2} & \dltp[18]{4.7} & \textbf{81.5} & \dltp[23]{6.7} \\
\bottomrule
\end{tabularx}
\caption{\textbf{\MethodName against the baselines} (trained at $p{=}0.4$). The $p{=}0.4$ columns evaluate each policy with its training user (RQ2), the others with a less masked user (RQ3). \textsc{GRPO} starts from the base model, for \texttt{Qwen3.5-2B} with the settings of \MethodName, for \texttt{Qwen3.5-4B} as in \autoref{tab:main}. Success (\%); bold: best per column.}
\label{tab:sg}
\end{table*}

\paragraph{Weak users cost every policy.} \autoref{tab:main} reports the cross-evaluation and \autoref{fig:maskcost} the loss of every policy when its user is masked. On \taub{}, masking costs the untrained 4B model 6.2 and 14.7 points at $p{=}0.2$ and $0.4$ (\autoref{tab:main_full}), and the strong-trained policy loses 12.7 points at $p{=}0.4$; on \texttt{ToolWOZ} the strong-trained policies lose 10.9 (4B) and 11.2 (2B) points. The same holds throughout training (\autoref{fig:training}). On \taub{} the $p{=}0.4$ arm trails the unmasked arm on 76 of 80 steps; on \texttt{ToolWOZ} the step-paired cost is $-$4.9/$-$9.9 points for \texttt{Qwen3.5-4B} at $p{=}0.2/0.4$, against $+$0.1$\pm$0.4 for a rerun of the unmasked arm, and it does not narrow over eight epochs. \texttt{Qwen3.5-2B} pays $-$6.2/$-$6.6 points on \texttt{ToolWOZ} (epochs 1--4, before all arms diverge) and about $-$5.5 points on \taub{}.

\paragraph{Strong- versus weak-user training.} On \taub{}, RL with the strong user lifts every column for the 4B model; the strong-trained policy is the best policy under the unmasked and the 0.2 evaluation, and at 0.4 no weak-trained policy beats it beyond noise. On \texttt{ToolWOZ} the 4B policies are within 2.5 points of each other on strong users, and the best weak-user scores (hybrid 77.8 and fixed 0.2 at 77.1 for 4B, fixed 0.2 at 66.8 for 2B) exceed the strong-trained policy by about 3 points, within the 6.1-point spread between three runs of one configuration (\autoref{tab:app_tw}). The only difference beyond two task-paired standard errors in favour of weak-user training is the 2B \texttt{ToolWOZ} policy trained at $p{=}0.2$ and evaluated at the same mask ($+$5.7 points); it lies within the run-to-run spread and falls to $+$3.0 at $p{=}0.4$.

\paragraph{Fixed masks.} On \taub{} (4B), we compare each weak-trained policy with the strong-trained one on the same tasks. Across the 24 fixed-mask comparisons of the full matrix (two masked arms $\times$ checkpoints at epochs 2, 4, 6, 8 $\times$ three evaluation masks), 22 are negative, the mean is $-$7.2 points, and all six differences beyond two standard errors are negative. Under the matching masked evaluation the weak-trained policies are \emph{not} better than the strong-trained one ($-$1.0 and $-$2.0 points at $p{=}0.4$), and they pay 7--9 points on unmasked users. Their smaller own masking cost in \autoref{tab:main} comes from a lower unmasked score, not from a higher masked one.

\paragraph{Mixtures and curricula.} Per-sample mixtures (hybrid) and a curriculum do not raise the weak-user score beyond noise either: at $p{=}0.4$ they recover 15\% and 8\% of the gap on \taub{} (4B) and 29\% and 7\% on \texttt{ToolWOZ} (4B), and for \texttt{Qwen3.5-2B} on \texttt{ToolWOZ} they lose 9--11 points on unmasked users (\autoref{app:mixtures}).

\paragraph{Where the loss comes from.} \autoref{tab:main} separates the test gap, $P(\trSW)-P(\trSS)$, from the training-condition gap, $P(\trWW)-P(\trSW)$. For 4B these are respectively $-$12.7 and $-$2.0 points on \taub{}, and $-$10.9 and $+$0.2 on \texttt{ToolWOZ}: weak-user training does not reliably offset the test-time loss. The 2B \taub{} results are noisier; neither gap exceeds two standard errors (\autoref{app:2b}).

\begin{figure}[t]
\centering
\includegraphics[width=\columnwidth]{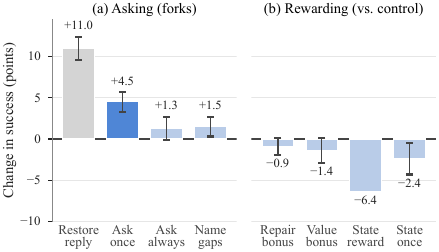}
\caption{\textbf{What does not help} (\texttt{ToolWOZ}; exploratory single runs). (a) Gain over the masked dialogue in counterfactual forks (all 150 prefixes; asking once recovers 25\% on the 105 audited ones). (b) Success minus an identical control for repair bonuses (4B) and state-accuracy rewards (2B). Bars: $\pm$1\,SE.}
\label{fig:nothelp}
\end{figure}

\paragraph{Asking more is not the answer.} If the agent lacks information, the obvious fix is to make it ask. On \texttt{ToolWOZ}, a prompt telling the agent that the customer's reply may be incomplete raises the rate of asking after a masked turn fivefold, but changes success at mask 0.4 by only $+$1.9 (untrained) and $-$1.2 points (strong-trained), against $+$16.8 for RL training. In counterfactual forks of 150 \texttt{ToolWOZ} dialogue prefixes where masking appeared to remove a task value, restoring the original reply gains $+$11.0$\pm$1.4 points; a single request to repeat at that point recovers 41\% of this (25\% on the 105 prefixes whose labels survive an audit; \autoref{fig:nothelp}a), asking whenever something is masked 12\%, and telling the agent which booking fields are missing 14\%. On \taub{}, when a user-stated identifier is masked, the trained agent already asks in 76\% of cases and 94\% of these gaps close, which accounts for only about one point of the 12.7-point loss. The agent does not lack questions; it lacks knowing which requirement of the task is still open (details: \autoref{app:ask}).

\paragraph{Neither does rewarding repairs or the state.} If the agent does not ask well, the reward might teach it. A bonus for repair moves, given for any repair or only for one that recovers a masked value, is too sparse to do so: only 2--7\% of groups receive a signal that differs from the task reward, and neither bonus changes held-out success (\autoref{fig:nothelp}b). Rewarding the state the agent writes for its accuracy fails for the opposite reason: within a group the state score is unrelated to success (correlation $-$0.07 to $+$0.04), so it adds noise to the task signal, and a state written once after the dialogue is learned without improving the task (\autoref{app:path}). A written state need not be correct to be useful, however: rollouts that write the same state are in comparable situations, and \MethodName uses it to compare their turns.

\subsection{\texorpdfstring{\rqtag{2}{rqtwo}\ }{RQ2: }Does \MethodName help with the user it is trained with?}
\label{sec:rq4}

\begin{table}[t]
\centering
\small
\setlength{\tabcolsep}{3.5pt}
\begin{tabular}{lccc}
\toprule
Trained and evaluated at & $p{=}0$ & 0.2 & 0.4 \\
\midrule
\grp{4}{Qwen3.5-2B} \\
\textsc{GRPO} & 74.9 & \textbf{74.8} & 72.0 \\
\textsc{UserRL} & 76.5 & 69.3 & 57.9 \\
\textsc{InfoPO} & \textbf{85.3} & 72.4 & 74.0 \\
\rowcolor{oursbg} \MethodName & 82.3\dltp{7.4} & \textbf{74.8}\dlt{$\pm$0.0} & \textbf{76.8}\dltp{4.8} \\
\midrule
\grp{4}{Qwen3.5-4B} \\
\textsc{GRPO} & 85.5 & 82.4 & 74.8 \\
\textsc{UserRL} & 86.1 & 48.7 & 59.1 \\
\textsc{InfoPO} & 87.9 & 82.1 & 77.9 \\
\rowcolor{oursbg} \MethodName & \textbf{88.1}\dltp{2.6} & \textbf{87.1}\dltp{4.7} & \textbf{81.5}\dltp{6.7} \\
% 4B GRPO row: from base, lr 3e-5; same-start 4B SFT,GRPO rows pending user confirmation
\bottomrule
\end{tabular}
\caption{\textbf{Training mask} (\texttt{ToolWOZ}). Each policy is trained and evaluated with the same user; \textsc{GRPO}: from the base model, at mask 0.4 for \texttt{Qwen3.5-2B} as in \autoref{tab:sg}, otherwise the policies of \autoref{tab:main}; small numbers: change from \textsc{GRPO}; bold: best per column. Standard deviations: \autoref{tab:sgmask_full}.}
\label{tab:sgmask}
\end{table}

% Forest plot of \MethodName minus GRPO (same information as the grey deltas of tab:sg; commented out)
\iffalse
\begin{figure}[!tb]
\centering
\includegraphics[width=\columnwidth]{figures/fig_sg_gain.pdf}
\caption{\textbf{\MethodName minus \textsc{GRPO} from the same cold start} ($p{=}0.4$, task-paired; thick $\pm$1\,SE, thin $\pm$2\,SE, filled beyond 2\,SE).}

\end{figure}
\fi

% Earlier single-column version (replaced by fig:why)
\iffalse
\begin{figure}[!tb]
\centering
\includegraphics[width=\columnwidth,trim=0 0 250 0,clip]{figures/figD_stability.pdf}
\caption{\textbf{\MethodName keeps improving while moving less from the reference policy} (\texttt{ToolWOZ}, SFT start, $p{=}0.4$).}

\end{figure}
\fi

\begin{figure}[t]
\centering
\includegraphics[width=0.95\columnwidth]{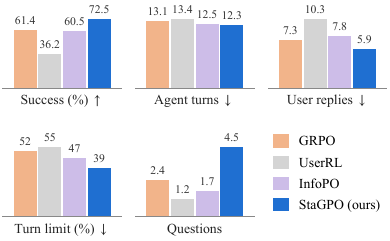}
\caption{\textbf{Interaction with the weak user} (\texttt{ToolWOZ}, \texttt{Qwen3.5-2B}, trained and evaluated at $p{=}0.4$; final checkpoint, step 120; per test dialogue). \textsc{GRPO} starts from the same SFT cold start as \MethodName. Turn limit: share of dialogues that use all 16 agent turns.}
% Turn limit: \% of dialogues that use all 16 agent turns; questions: agent messages that ask the user something.}
\label{fig:user}
\end{figure}

\paragraph{Against the baselines.} \MethodName reaches 76.8 with weak users on \texttt{ToolWOZ}, 4.8 points above \textsc{GRPO} (72.0), 2.8 above \textsc{InfoPO} (74.0) and 18.9 above \textsc{UserRL} (\autoref{tab:sg}, $p{=}0.4$). Each of its three runs (75.7--77.4; \autoref{tab:app_tw}) is above every baseline. Mixing masks during training does not close the gap: the hybrid and curriculum policies reach 52.0 and 62.9. On \texttt{Qwen3.5-4B}, \MethodName reaches 81.5, 6.7 points above \textsc{GRPO} (74.8) and 3.6 above \textsc{InfoPO} (77.9).

\paragraph{Across training masks.} In \autoref{tab:sgmask}, \MethodName is 7.4 points above \textsc{GRPO} when both are trained with the strong user (82.3 vs.\ 74.9), level at mask 0.2 (74.8 vs.\ 74.8) and 4.8 points above at 0.4 (76.8 vs.\ 72.0). On \texttt{Qwen3.5-4B} it is the best policy at every mask, 2.6, 4.7 and 6.7 points above \textsc{GRPO}. The exception is \textsc{InfoPO} trained and evaluated with the strong user, which is 3.0 points above \MethodName for \texttt{Qwen3.5-2B} (85.3 vs.\ 82.3).

\paragraph{\taub{}.} With weak users, \MethodName is the best \texttt{Qwen3.5-2B} policy on \taub{}, 16.6 points above \textsc{GRPO} and 7.3 above the next-best baseline (curriculum, 33.3; \autoref{tab:sg}, $p{=}0.4$). It also scores highest for \texttt{Qwen3.5-4B}, 3.5 points above \textsc{GRPO}, though only 1.0 above the hybrid policy (41.7), within evaluation noise. \MethodName uses a shaped reward on this benchmark (\autoref{app:hparams}).

\begin{figure}[t]
\centering
\includegraphics[width=0.9\columnwidth]{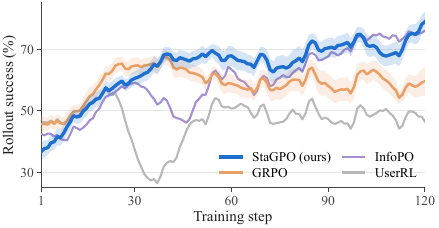}
\caption{\textbf{Training with the weak user} (\texttt{ToolWOZ}, \texttt{Qwen3.5-2B}, $p{=}0.4$): rollout success per step, rolling mean over 13 steps; shading: $\pm$1\,SE. \textsc{GRPO} here starts from the same SFT cold start as \MethodName.}
\label{fig:curves2b}
\end{figure}

\paragraph{How training differs.} On \texttt{ToolWOZ}, \MethodName achieves the highest mean training success (63.8\%, versus 59.6\% for \textsc{InfoPO}, 58.9\% for \textsc{GRPO} from the same SFT start and 46.8\% for \textsc{UserRL}) and finishes at 79\% (\autoref{fig:curves2b}). At the final checkpoint (\autoref{fig:user}), it needs fewer user replies than the baselines shown there (5.9 versus 7.3--10.3 per dialogue) and reaches the turn limit less often (39\% versus 47--55\%), despite asking more questions, while its KL is about a quarter of that of \textsc{GRPO} from the same cold start (\autoref{app:sgfigs}). With the SFT cold start, the share of anchored turns rises from 30\% in the first 20 steps to 48\% in the last 20; without it, it falls from 24\% to 19\% (\autoref{tab:cold}). \MethodName also beats \textsc{GRPO} with the strong user (\autoref{tab:sgmask}): it is a better multi-turn credit assignment in general. On \texttt{ToolWOZ} (2B, $p{=}0.4$), grouping, not the written state, is what helps: from the same cold start, \textsc{GRPO} that writes the same states reaches 60.1 and 69.0 with weak and strong users, 16.7 and 15.1 points below \MethodName and no better than without the state (62.5; \autoref{tab:ablation}).

\subsection{\texorpdfstring{\rqtag{3}{rqthree}\ }{RQ3: }Do the gains transfer across users?}
\label{sec:rq5}

\begin{figure}[t]
\centering
\includegraphics[width=0.95\columnwidth]{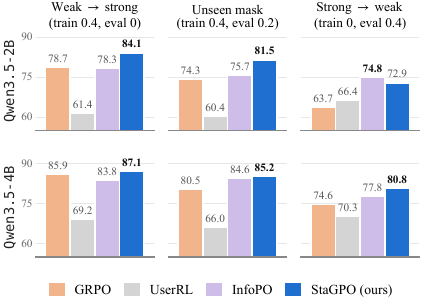}
\caption{\textbf{Transfer across users} (\texttt{ToolWOZ}, test success, \%). Each policy is evaluated with a user it was not trained with; bold: best per setting.}
\label{fig:transfer}
\end{figure}

\autoref{fig:transfer} evaluates every policy on \texttt{ToolWOZ} with a user it was not trained with.

\paragraph{From weak to strong users.} Trained with weak users and evaluated with strong ones, \MethodName scores 84.1 (2B) and 87.1 (4B), the highest of all policies trained at $p{=}0.4$, 5.4 and 1.2 points above \textsc{GRPO}. For \texttt{Qwen3.5-2B} it is on par with \MethodName trained with the strong user itself (82.3): the incomplete weak-user dialogues teach a policy that also serves strong users.

\paragraph{To an unseen mask rate.} Trained at mask 0.4 and evaluated at 0.2, a rate it never saw, \MethodName is again the best policy for both models (81.5 and 85.2), 7.2 and 4.7 points above \textsc{GRPO}; \textsc{InfoPO} comes close for \texttt{Qwen3.5-4B} (84.6).

\paragraph{From strong to weak users.} This is the setting of RQ1, in which the strong-trained \textsc{GRPO} policy loses 11.2 (2B) and 10.9 (4B) points when its user is masked at 0.4. Trained only with strong users, \MethodName serves weak users at 72.9 (2B) and 80.8 (4B), 9.2 and 6.2 points above strong-trained \textsc{GRPO}, and level with (2B: 72.0) or above (4B: 74.8) \textsc{GRPO} trained with weak users. For \texttt{Qwen3.5-4B} it is the best strong-trained policy; for \texttt{Qwen3.5-2B}, \textsc{InfoPO} is higher (74.8).

\paragraph{\taub{}.} The $p{=}0$ and 0.2 columns of \autoref{tab:sg} give the same transfer on \taub{} for policies trained at 0.4. \MethodName stays above \textsc{GRPO} for both models (2B: $+$11.4 and $+$6.3; 4B: $+$3.9 and $+$2.6) and is the best or tied-best policy at $p{=}0$ (2B: level with the curriculum policy at 45.8; 4B: 51.0), but at $p{=}0.2$ \textsc{InfoPO} (2B) and the hybrid policy (4B) are higher. The evidence for transfer therefore rests mainly on \texttt{ToolWOZ}, where the 4B results are single runs.

\paragraph{What transfers.} Gains transfer across masking conditions on \texttt{ToolWOZ}; other simulators and naturally incomplete users remain untested.

\section{Conclusion}

Standard weak-user training does not consistently recover the loss from incomplete information. \MethodName groups agent-written need--gap--fill states for step-level credit assignment. With an SFT cold start, it improves weak-user performance and transfers across user conditions, most clearly on \texttt{ToolWOZ}.

\section*{Limitations}

Our weak users are simulated by removing sentences from an otherwise cooperative user. This isolates one way in which users can be weak; real users may also be vague, inconsistent or mistaken, and studying these behaviours is a natural next step. Our experiments cover two benchmarks, two model sizes and one model family; larger models, other domains and human users may behave differently. Because the user simulator cannot be seeded, evaluations carry sampling noise, which we address with task-paired standard errors and repeated evaluations; due to compute, several configurations were trained once. The benchmark rewards measure task completion and do not capture every aspect of interaction quality, such as whether the agent confirms before acting. Masked sentences are replaced by an explicit \texttt{[inaudible]} marker; users who leave information out leave no such trace, and silent omissions are a natural extension. Finally, the abstraction that maps a written state to an anchor key is designed per domain, and learning it automatically is left to future work.

\section*{Ethics Statement}

All users in this work are simulated; no human subjects were involved. Our motivation is to make agents serve users who describe their needs less completely, who are often those with the least technical experience, as language models increasingly mediate access to services and public information \citep{bao2026policyllm}. Agents that act without confirmation, which we observe after training, could harm such users in deployed settings; interfaces that make model behaviour visible can help users calibrate their trust \citep{bao2026intraai}.

\bibliography{custom}

\appendix
\appendixautoref

\clearpage
\section{Experimental Details}
\label{app:train}

\subsection{Environments}
\label{app:env}

All \taub{} experiments use its retail domain (\taur{}), in which the user has no tools of its own and so cannot complete the task for the agent; we chose it so that success reflects the agent's actions rather than the simulated user's. The 80 training tasks and the 34-task evaluation pool are drawn from this domain. The evaluation pool consists of 10 development and 24 previously used test tasks; the user is \texttt{gpt-oss-120b} with a budget of 45 agent messages, and the reward is the benchmark's binary database check. \texttt{ToolWOZ} uses booking tasks over restaurant, hotel, train and attraction APIs, with 120 training tasks; all main experiments use \texttt{gpt-oss-120b} as the user, and \texttt{gpt-4o-mini} and an earlier 21-task test split appear only in the exploratory experiments of \autoref{app:path}. For the \MethodName study, 100 further \texttt{ToolWOZ} tasks disjoint from the training data are split into 30 validation and 70 test tasks, with a locally served \texttt{gpt-oss-120b} as the user; on \taur{}, the 10 development tasks serve for validation and the 24 test tasks for evaluation.

\subsection{Training}
\label{app:hparams}

We train \texttt{Qwen3.5-4B} and \texttt{Qwen3.5-2B} in non-thinking mode with \textsc{GRPO} in verl \citep{sheng2024hybridflow} and LoRA \citep{hu2022lora}, with a KL coefficient of 0.005. On \taur{} we use LoRA rank 32, 8 tasks $\times$ 8 rollouts per step and learning rate $3{\times}10^{-5}$ for 8 epochs (80 steps); on \texttt{ToolWOZ}, 8 tasks per step and learning rate $3{\times}10^{-5}$ for 8 epochs (120 steps). All arms of a benchmark share the task order, so the $n$-th step of every arm sees the same tasks. Besides fixed masks we train two schedules with the same average rate of 0.2: \emph{hybrid}, where training samples are assigned masks 0, 0.2 and 0.4 in rotation, and \emph{curriculum}, where the mask grows linearly from 0 in epoch~1 to 0.4 in epoch~8. In the \MethodName study all arms use LoRA rank 32, a KL coefficient of 0.005, clipping bounds $\varepsilon_{\text{low}}{=}0.2$ and $\varepsilon_{\text{high}}{=}0.28$, 8 tasks $\times$ 8 rollouts per step and \texttt{gpt-oss-120b} as the user; \texttt{ToolWOZ} dialogues are limited to 16 agent turns and \taur{} trajectories to 8,192 tokens. \textsc{InfoPO} and \textsc{UserRL} replace only the advantage computation and are otherwise identical to the corresponding \textsc{GRPO} run. On \taur{} the state follows a compact retail schema, and the anchor key of \autoref{eq:key} consists of the user's intent, the identity-verification status, the order and items in focus, the missing fields, the last result and whether the request is done, together with the name of the last tool call and the number of consecutive tool errors (0, 1, 2+); the step advantage of \autoref{eq:step} is added to the action tokens only.

\paragraph{Cold start and baselines.} The SFT cold start uses 33 of the model's own training dialogues and no stronger model (\textsc{UserRL} distils about 5K trajectories with \texttt{GPT-4o}); the agent writes the state from the instruction alone. The \MethodName study trains at mask 0.4 unless noted, for 120 steps at learning rate $3{\times}10^{-5}$ on \texttt{ToolWOZ} and 80 steps at $3{\times}10^{-5}$ on \taur{}. The \textsc{GRPO} baseline of \texttt{Qwen3.5-2B} is trained from the base model with these settings (learning rate, steps, checkpoint selection) but without the cold start; the \textsc{GRPO} baseline of \texttt{Qwen3.5-4B} and the hybrid and curriculum baselines are the policies of the cross-evaluation (\autoref{tab:main}). Further baselines are in \autoref{app:sgfull}, and \MethodName without the cold start in \autoref{app:nocold}. The shaped reward on \taur{} gives a failed dialogue 0.5 times the fraction of correct write actions, and every tool call that returns an error costs 0.05 (at most 0.5).

\subsection{Choosing a Second Model}
\label{app:screen}

We screened candidate models untrained, under the same three evaluation masks. \texttt{Qwen3.5-9B} (45.4/37.6/37.9) and \texttt{Qwen3.5-2B} (41.2/33.0/30.7) show significant masking costs. \texttt{Qwen3-8B} succeeds on only about 15\% of tasks with no significant masking cost, and \texttt{Llama-3.2-3B} stays near zero success on \taur{} because it loops on tool calls instead of talking to the user; both would make masking uninformative. We chose \texttt{Qwen3.5-2B} for cost.

\subsection{Channel Masking}
\label{app:mask}

Each user reply is split into sentences at sentence-final punctuation followed by whitespace, or at a line break. A sentence is replaced by the token \texttt{[inaudible]} when a uniform draw falls below the mask rate $p$; the surrounding whitespace is kept, so the agent sees where something was lost but not what. On \taub{}, sentences that carry the control markers ending an episode (stop, transfer, out of scope) are never masked but count towards the number of sentences, which is why the realised rate is slightly below $p$. The draw is a deterministic hash of the task, the index of the user turn and the index of the sentence (on \taub{} also an evaluation seed, fixed during training), so a lower mask rate masks a subset of the sentences masked at a higher rate, and on \texttt{ToolWOZ} every rollout and evaluation of a task loses the same sentences. The opening user message is never masked. The simulator is unchanged and always sees its own unmasked history. A constructed example at $p{=}0.4$:

\begin{quote}\small
\textit{Said:} I'd like to return the water bottle from my last order. The order number is \#W2378156. Please refund it to the original credit card.\\[2pt]
\textit{Heard:} I'd like to return the water bottle from my last order. \texttt{[inaudible]} Please refund it to the original credit card.
\end{quote}

\noindent The agent can recover the order number only by asking for it, or by looking it up through the user's account where the domain allows it.

\subsection{State Format and Anchor Groups}
\label{app:state}

Before every action the agent writes a JSON object between \texttt{<state>} tags; the block is removed before the reply reaches the user simulator. On \texttt{ToolWOZ} the object has seven fields: \texttt{goal}, \texttt{pending\_task}, \texttt{requirements} (the booking slots and the values known so far), \texttt{candidates} (entities returned by search), \texttt{selected\_entity}, \texttt{missing\_user\_information}, and \texttt{attempts}, the last three tool calls written as \texttt{name(arguments) -> outcome} with an outcome of $N$ results, empty, booked or failed. The first five fields form the need $n$, \texttt{missing\_user\_information} the gap $g$, and \texttt{attempts} the fill $f$. The format is not taught by supervision: the SFT cold start uses 33 of the model's own successful training dialogues without any state (dialogues with an exactly repeated tool call are removed), and the state is requested only by an instruction in the system prompt during RL, which the SFT model follows with all seven fields in 79\% of turns before training.

The anchor key (\autoref{eq:key}) is the pending domain, the pending action, the set of requirement slots whose value is known, and the outcome type of the last tool call (results, empty, failed, booked or none). Slot values are not part of the key, so two rollouts that both know a date share a key even if the dates differ. A group is kept if its turns come from at least $m{=}3$ of the $G{=}8$ rollouts; turns whose state cannot be parsed receive no key. On \texttt{ToolWOZ} the step advantage, clipped to $\pm3$, is added to every token of the turn (state and action); on \taub{} only to the action tokens (\autoref{app:train}). Anchoring requires only that rollouts in the same situation write the same key, not that the state is correct: a gap that is written wrongly but consistently still groups comparable turns.

\autoref{tab:anchor_example} works through a constructed group. Rollouts 1 and 3 ask for the missing day and later book successfully; rollout 2 books without it, fails and runs out of turns. With four of the eight rollouts successful, $\sigma_q{=}0.5$ and the episode advantages are $\pm1$; the step advantage separates the two successful rollouts by how soon they finish and adds a further penalty to the turn that led to failure.

\begin{table}[h]
\centering
\small
\setlength{\tabcolsep}{4pt}
\begin{tabular}{lccccc}
\toprule
Rollout & $t$ / $T$ & $R$ & $G_{i,t}$ & $A^{\text{step}}$ & $A$ \\
\midrule
1: asks for the day & 4 / 9 & 1 & 0.774 & $+$0.46 & $+$1.46 \\
2: books without it & 5 / 16 & 0 & 0 & $-$1.09 & $-$2.09 \\
3: asks for the day & 4 / 7 & 1 & 0.857 & $+$0.63 & $+$1.63 \\
\midrule
Group mean $\bar G_{\mathcal{A}}$ & & & 0.544 & & \\
\bottomrule
\end{tabular}
\caption{A constructed anchor group with key (restaurant, book, \{area, food, people\}, results): three rollouts reach it at turn $t$ of $T$; $\gamma{=}0.95$, $\sigma_q{=}0.5$, $\omega{=}1$.}
\label{tab:anchor_example}
\end{table}

\subsection{Statistical Protocol}
\label{app:stats}

The unit of analysis is the task. Each policy is evaluated on every test task several times, and a task's score is its success rate over these repetitions; where a checkpoint was evaluated twice, the repetitions are pooled. Two policies are compared through task-paired differences $d_q{=}s_A(q)-s_B(q)$ over the same tasks, reported as the mean and its standard error $\mathrm{sd}(d)/\sqrt{n}$ over the $n$ tasks; a difference is bold when it exceeds two standard errors. These standard errors reflect the choice of tasks and the sampling of the simulator, not the variation between training runs. Where a configuration has several training runs, tables report the mean over runs and \autoref{tab:app_tw} and \autoref{tab:app_rt} list every run. Standard deviations in \autoref{tab:main_full}, \autoref{tab:sg_full} and \autoref{tab:sgmask_full} are over evaluation rounds, where a round is one evaluation of one training run; for configurations with several runs they pool the rounds of all runs.

Three evaluations of the same untrained \texttt{Qwen3.5-4B} model span 4.9 points in the worst cell. The recovery rate $\rho(M)=\frac{P(M\to\textsc{W})-P(\trSW)}{P(\trSS)-P(\trSW)}$ gives the share of the strong-trained policy's weak-user loss that a training method $M$ recovers. For the \texttt{Qwen3.5-4B} hybrid policy on \texttt{ToolWOZ} at $p{=}0.4$ it is $(77.8-74.6)/(85.5-74.6)=29\%$.

In the cross-evaluation (\autoref{tab:main}) every checkpoint is evaluated under every mask: \taub{} policies are taken after 8 epochs (the 2B hybrid policy at step 60, before its training collapsed) and evaluated on the 34 tasks $\times$ 3 seeds, \texttt{ToolWOZ} policies (one run per arm; \texttt{Qwen3.5-2B} at learning rate $3{\times}10^{-5}$) on the 70 test tasks $\times$ 2. In the \MethodName study, \texttt{ToolWOZ} policies are evaluated on the 70 test tasks $\times$ 2 and \taub{} policies on the 24 test tasks $\times$ 4 seeds; in \autoref{tab:sg}, the untrained and \texttt{Qwen3.5-4B} \textsc{GRPO} rows on \taub{} are the cross-evaluation results above. \texttt{Qwen3.5-4B} rows have one run each. Checkpoint selection uses validation scores. The candidate steps are fixed before training (30/60/90/120 on \texttt{ToolWOZ}, 20/40/60/80 on \taub{}); the step with the highest validation score under the training mask is selected, and only its test scores are reported. On \taub{}, \textsc{UserRL} (2B), whose training stopped at step 40, is compared at steps 20 and 40, and the \texttt{Qwen3.5-4B} runs at the steps listed in \autoref{tab:app_rt}.

\FloatBarrier
\section{Exploratory Study: Simulator Scale}
\label{app:scale}

A natural way to obtain stronger and weaker users is to change the size of the simulator. In an exploratory study run before the main protocol was fixed, we trained \texttt{Qwen3.5-4B} from the base model with \textsc{GRPO} on \taur{}, with either \texttt{gpt-oss-20b} or \texttt{gpt-oss-120b} as the user and the agent, tasks and user persona fixed (16 tasks $\times$ 8 rollouts per step, 40 steps; checkpoints after epochs 4 and 8), and evaluated every policy with both simulators on the 34 evaluation tasks, three times each (\autoref{tab:scale}).

\begin{table}[h]
\centering
\small
\setlength{\tabcolsep}{4pt}
\begin{tabular}{lccc}
\toprule
 & \multicolumn{2}{c}{Evaluation user} & \\
\cmidrule(lr){2-3}
Training user (epoch) & \texttt{20b} & \texttt{120b} & Mean \\
\midrule
none (untrained) & 35.3\sd{7.8} & 39.2\sd{7.4} & 37.3 \\
\texttt{gpt-oss-20b} (4) & 41.2\sd{2.9} & 36.3\sd{9.5} & 38.7 \\
\texttt{gpt-oss-120b} (4) & 37.3\sd{6.8} & 45.1\sd{7.4} & 41.2 \\
\texttt{gpt-oss-20b} (8) & 43.1\sd{4.5} & 44.1\sd{2.9} & 43.6 \\
\texttt{gpt-oss-120b} (8) & 46.1\sd{3.4} & 44.1\sd{5.9} & 45.1 \\
\bottomrule
\end{tabular}
\caption{\textbf{Simulator scale} (\texttt{Qwen3.5-4B}, \taur{}, 34 tasks; success, \%, $\pm$ standard deviation over three evaluation rounds; single training runs).}
\label{tab:scale}
\end{table}

After four epochs, each policy does better with the simulator it was trained with: comparing the two policies under the same evaluation user, the 20b-trained policy is 3.9 points higher with \texttt{gpt-oss-20b} and the 120b-trained policy 8.8 points higher with \texttt{gpt-oss-120b}. After eight epochs this ordering is gone: the 120b-trained policy is higher with \texttt{gpt-oss-20b}, and both score 44.1 with \texttt{gpt-oss-120b}; averaged over both evaluation users, the 120b-trained policy is 1.5 points higher. These differences are within the standard deviations over evaluation rounds (2.9--9.5 points), and each arm was trained once, so we do not read them as a stable advantage of either simulator. They show only that simulator size does not induce a fixed ordering of users: which simulator is the better training partner depends on the evaluation user and on when training stops, and a different simulator changes several properties of the interaction at once. We therefore keep the simulator fixed and control directly how much of the user's information reaches the agent.

\FloatBarrier
\section{Further Results for RQ1}
\label{app:xeval}

\subsection{Full Cross-Evaluation}
\label{app:fullx}

% Heatmap of weak- minus strong-trained policies (same numbers as tab:main; removed)
\iffalse
\begin{figure}[t]
\centering
\includegraphics[width=\columnwidth]{figures/fig_xeval_v3.pdf}
\caption{\textbf{Weak-trained minus strong-trained policies.} Task-paired difference in success (points) between each policy trained with a fixed mask and the strong-trained policy of the same model and benchmark (\autoref{tab:main}), by evaluation mask; bold: beyond two standard errors.}
\label{fig:xeval}
\end{figure}
\fi

\paragraph{Mixtures and curricula.}\label{app:mixtures} On \taub{} (4B) the hybrid and curriculum schedules keep the unmasked score (within 3 points of the strong-trained policy) but do not improve the masked one: $+$2.0 and $+$1.0 points at $p{=}0.4$, recovery rates of 15\% and 8\%, both within noise. The curriculum's own masking cost ($-$10.8) remains significant. On \texttt{ToolWOZ} (4B) the hybrid policy recovers 29\% of the gap ($-$8.3 against $-$10.9 for the strong-trained policy) and the curriculum 7\% ($-$11.1), from single runs. For \texttt{Qwen3.5-2B} on \texttt{ToolWOZ} neither schedule beats the strong-trained policy under the weak user (52.0 and 62.9 against 63.7), and both lose 9--11 points on unmasked users (64.0 and 65.7 against 74.9). On \taub{} the 2B hybrid policy, taken at step 60, ties the strong-trained policy under the weak user (33.3 vs.\ 33.3) and is 2.9 points lower on unmasked users (37.3 vs.\ 40.2). Standard deviations are in \autoref{tab:main_full}.

\begin{table*}[t]
\centering
\setlength{\tabcolsep}{2pt}
\begin{tabularx}{\textwidth}{ll*{6}{>{\centering\arraybackslash}X}}
\toprule
 & & \multicolumn{3}{c}{\taub{}} & \multicolumn{3}{c}{\texttt{ToolWOZ}} \\
\cmidrule(lr){3-5}\cmidrule(lr){6-8}
Type & Method & $p{=}0$ & 0.2 & 0.4 & $p{=}0$ & 0.2 & 0.4 \\
\midrule
\grp{8}{Qwen3.5-4B} \\
Prompting & Untrained & 46.7\sd{4.3} & 40.5\sd{4.6} & 32.0\sd{5.6} & 77.3\sd{2.7} & 72.6\sd{6.1} & 69.0\sd{4.6} \\
RL, strong user & \textsc{GRPO} ($p{=}0$) & 53.9\sd{3.4} & 51.0\sd{10.3} & 41.2\sd{8.8} & 85.5\sd{2.9} & 80.5\sd{1.1} & 74.6\sd{2.4} \\
RL, weak user & \textsc{GRPO} ($p{=}0.2$) & 45.1\sd{10.3} & 39.2\sd{6.1} & 40.2\sd{3.4} & 84.0\sd{0.2} & 82.4\sd{0.6} & 77.1\sd{1.4} \\
RL, weak user & \textsc{GRPO} ($p{=}0.4$) & 47.1\sd{10.6} & 42.2\sd{3.4} & 39.2\sd{1.7} & 85.9\sd{2.1} & 80.5\sd{0.3} & 74.8\sd{2.7} \\
RL, weak user & Hybrid (0/0.2/0.4) & 51.0\sd{4.5} & 44.1\sd{7.8} & 43.1\sd{10.3} & 86.1\sd{2.0} & 82.7\sd{0.9} & 77.8\sd{0.1} \\
RL, weak user & Curriculum (0$\to$0.4) & 52.9\sd{5.9} & 44.1\sd{5.1} & 42.2\sd{7.4} & 86.5\sd{2.1} & 80.1\sd{4.8} & 75.4\sd{0.1} \\
\midrule
\grp{8}{Qwen3.5-2B} \\
Prompting & Untrained & 41.2\sd{5.3} & 33.0\sd{8.2} & 30.7\sd{4.9} & 49.0\sd{2.4} & 42.6\sd{1.8} & 40.7\sd{3.9} \\
RL, strong user & \textsc{GRPO} ($p{=}0$) & 40.2\sd{8.3} & 36.6\sd{10.0} & 33.3\sd{7.8} & 74.9\sd{0.2} & 69.2\sd{2.0} & 63.7\sd{0.1} \\
RL, weak user & \textsc{GRPO} ($p{=}0.2$) & 34.3\sd{4.5} & 29.4\sd{8.8} & 26.5\sd{5.9} & 76.1\sd{0.3} & 74.8\sd{6.3} & 66.8\sd{2.7} \\
RL, weak user & \textsc{GRPO} ($p{=}0.4$) & 44.4\sd{7.8} & 40.5\sd{5.3} & 38.9\sd{6.5} & 71.2\sd{0.1} & 67.7\sd{2.4} & 62.1\sd{2.3} \\
RL, weak user & Hybrid (0/0.2/0.4) & 37.3\sd{11.1} & 39.2\sd{9.5} & 33.3\sd{7.4} & 64.0\sd{5.5} & 57.1\sd{0.6} & 52.0\sd{2.2} \\
RL, weak user & Curriculum (0$\to$0.4) & 51.0\sd{4.5} & 35.3\sd{10.2} & 34.3\sd{4.5} & 65.7\sd{2.4} & 65.1\sd{1.9} & 62.9\sd{4.9} \\
\bottomrule
\end{tabularx}
\caption{\textbf{Full cross-evaluation} (\autoref{tab:main} with standard deviations). Success (\%) $\pm$ standard deviation over evaluation rounds. Columns: the user's mask at evaluation ($p{=}0$: strong); rows: the user at training.}
\label{tab:main_full}
\end{table*}

\subsection{Training Curves on \texorpdfstring{$\tau^2$}{tau2}-retail}
\label{app:curves}

\begin{figure}[h]
\centering
\includegraphics[width=\columnwidth]{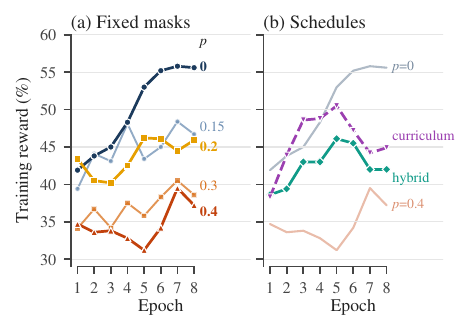}
\caption{Mean training reward per epoch, \texttt{Qwen3.5-4B} on \taur{}, each arm under its own mask.}
\label{fig:curves}
\end{figure}

\autoref{fig:curves} gives the mean training reward per epoch. Over the last four epochs the fixed-mask arms average 54.9, 45.9, 45.7, 38.3 and 35.5 at masks 0, 0.15, 0.2, 0.3 and 0.4, the hybrid arm 43.9 and the curriculum 46.7. The hybrid arm rotates masks 0, 0.2 and 0.4 across samples; the curriculum mask rises from 0 in epoch~1 to 0.4 in epoch~8, so its later epochs are trained under heavier masking.

% Table version of the same data (replaced by fig:curves, kept for reference)
\iffalse
\begin{table}[h]
\centering
\small
\setlength{\tabcolsep}{3pt}
\begin{tabular}{lcccccccc}
\toprule
Mask & E1 & E2 & E3 & E4 & E5 & E6 & E7 & E8 \\
\midrule
0 & 41.9 & 43.8 & 45.0 & 48.3 & 53.0 & 55.2 & 55.8 & 55.6 \\
0.15 & 39.4 & 44.1 & 43.1 & 48.1 & 43.4 & 45.0 & 48.4 & 46.7 \\
0.2 & 43.4 & 40.5 & 40.2 & 42.5 & 46.2 & 46.1 & 44.5 & 45.9 \\
0.3 & 34.1 & 36.7 & 34.2 & 37.5 & 35.8 & 38.3 & 40.5 & 38.6 \\
0.4 & 34.7 & 33.6 & 33.8 & 32.8 & 31.2 & 34.2 & 39.5 & 37.2 \\
\midrule
\textsc{Hybrid} & 38.6 & 39.4 & 43.0 & 43.0 & 46.1 & 45.5 & 42.0 & 42.0 \\
Curric. & 38.3 & 43.9 & 48.6 & 48.8 & 50.5 & 47.2 & 44.2 & 44.9 \\
\bottomrule
\end{tabular}
\caption{Mean training reward per epoch (\%), \texttt{Qwen3.5-4B} on \taur{}, each arm under its own mask.}

\end{table}
\fi

\subsection{Qwen3.5-2B on \texorpdfstring{$\tau^2$}{tau2}-retail}
\label{app:2b}

\autoref{tab:2b} pools three evaluations for the strong-trained and the $p{=}0.4$ policies; a single evaluation of the strong-trained policy gave 46.1 on unmasked users and two further evaluations gave 37.3 each. The first evaluation alone suggested a flat masking profile for the $p{=}0.4$ policy and a large lead of the strong-trained policy on unmasked users; both disappeared with more evaluations, which is why we pool.

\begin{table}[h]
\centering
\small
\setlength{\tabcolsep}{4pt}
\begin{tabular}{lccc}
\toprule
Policy & Eval 0 & Eval 0.2 & Eval 0.4 \\
\midrule
Untrained ($\times$3) & 41.2 & 33.0 & 30.7 \\
Strong-trained ($\times$3) & 40.2 & 36.6 & 33.3 \\
Fixed $p{=}0.2$ ($\times$1) & 34.3 & 29.4 & 26.5 \\
Fixed $p{=}0.4$ ($\times$3) & 44.4 & 40.5 & 38.9 \\
\midrule
0.4 $-$ strong-trained & $+$4.2 & $+$3.9 & $+$5.6 \\
0.4 $-$ untrained & $+$3.3 & \sig{$+$7.5} & \sig{$+$8.2} \\
\bottomrule
\end{tabular}
\caption{Qwen3.5-2B on \taur{}, epoch-8 checkpoints; ($\times k$): evaluations pooled.}
\label{tab:2b}
\end{table}

\FloatBarrier
\subsection{Side Effects of Training}
\label{app:side}

First, the \taur{} reward only checks the final database, and strong-user training teaches the agent to skip confirmation: when the user's ``yes'' is masked, the untrained 4B agent asks again 74\% of the time and proceeds 18\%, while the strong-trained agent proceeds 78\% of the time. Second, masked training changes how much the user says: on unmasked evaluation, dialogues of the $p{=}0.4$ policy contain 28.7 user sentences against 19.1 for the strong-trained policy, with no more user turns. Weak-trained agents buy robustness through redundancy, which costs efficiency and, on 4B, unmasked success.

\FloatBarrier
\subsection{Prompting, Counterfactual Forks and Lost Values}
\label{app:ask}
\label{sec:mechanism}
\label{app:audit}

\paragraph{Prompting does not help.} On \texttt{ToolWOZ}, adding ``the customer's reply may be incomplete; ask them to repeat before calling an API'' changes success at mask 0.4 by $+$1.9 points for the untrained model and $-$1.2 for the strong-trained one, both within noise, while training itself contributes $+$16.8. The prompt raises the rate of asking after a masked turn fivefold without changing the score.

\paragraph{Counterfactual forks.} On \texttt{ToolWOZ} we select 150 dialogue prefixes where the masking procedure appeared to remove a previously unseen task value, copy each state, and continue it four times per intervention. Against a within-batch masked control, restoring the original reply improves reward by $+$11.0$\pm$1.4 points. Forcing the agent to ask the user to repeat \emph{once} recovers $+$4.5$\pm$1.2 (41\%; \autoref{fig:nothelp}a); asking whenever something is masked recovers only $+$1.3$\pm$1.4, and telling the agent which booking fields are missing recovers $+$1.5$\pm$1.2.

\paragraph{What the lost information is.} On \taub{} we can follow every user-stated gold value (order ids, e-mails, zip codes, names, addresses) through the channel. In the 5,210 training dialogues of the $p{=}0.4$ arm, when such a value is masked, the agent asks in 76\% of cases, and 94\% of these gaps close, mostly because the user restates the value. Closed gaps cost 1.4 extra turns but no reward ($+$1.6$\pm$3.1 points within task); the 2.8\% of dialogues with a gap that never closes lose 28.9 points. This identifier-loss pathway accounts for only about one point of the 12.7-point test gap. Training also shapes behaviour the reward does not see: strong-user training teaches the agent to proceed when the user's confirmation is lost (\autoref{app:side}). Neither more questions nor more exposure to weak users teaches the agent which requirement of the task is still open.

\paragraph{Fork-label audit.} The initial selection script aligned the original and masked replies by sentence. Consecutive \texttt{[inaudible]} markers changed the sentence count and misaligned later labels: 39 of the 150 selected prefixes had no required booking parameter masked, and six more were partly mislabeled. This does not alter the observed rewards of the restore, mask, and ask-once branches, but it invalidates analyses that treat all selected prefixes as missing a booking parameter. In the 105 correctly labeled prefixes, restoring the reply gains $+$11.8 points and asking once gains $+$3.0 (25\% recovery); in the 39 without a masked required booking parameter, the respective gains are $+$7.7 and $+$7.2 (94\%). The latter group may contain lost search constraints or preferences. A booking-only intervention would therefore test only one part of the observed gap. We leave analyses based on the original ``criticality'' labels out of the main claim pending relabeling.

\paragraph{A limited missing-slot probe.} Asked which booking details the customer has not provided yet, the 4B policies flag 62--65\% of details marked missing by the original labels but also 53--58\% of details that were received; training does not change this. \texttt{gpt-oss-120b}, under the identical probe, flags 75\% and 2\%. Because the fork labels required correction and a prompted JSON probe need not reflect normal policy behavior, these figures are suggestive rather than proof that recognition is the sole bottleneck. The weak result from explicitly naming missing fields further shows that recognition alone need not produce an effective next action.

\FloatBarrier
\section{From Diagnosis to \MethodName}
\label{app:path}

The experiments in this section were run while the method was developed, before the protocol of the main experiments was fixed: \texttt{Qwen3.5-2B} on \texttt{ToolWOZ} at mask 0.4 unless noted, single runs, and earlier user simulators and evaluation protocols as stated. Their numbers are not comparable with the main tables; we report them as the observations that led to \MethodName, each with the hypothesis it rules out.

\paragraph{Asking more.} Prompting the agent to ask raises the rate of asking after a masked turn fivefold without changing success, and in counterfactual forks asking whenever something is masked recovers 12\% of what restoring the reply gains, against 41\% for a single request at the right point (\autoref{app:ask}). A lack of questions is not the bottleneck.

\paragraph{Recognising what is missing.} Asked which booking details the customer has not provided yet, the 4B policies also flag 53--58\% of the details they have received, and telling the agent which fields are missing recovers only 14\% of the gap in the forks (\autoref{app:ask}). The agent does not reliably know what is missing, and knowing it is not enough to act on it.

\paragraph{Rewarding repairs.} We added a bonus of 0.1 to the task reward either for any repair move, such as asking for missing information (A), or only for a move that recovers a masked value (B), and continued training for 40 steps from a strong-trained \textsc{GRPO} policy (\texttt{Qwen3.5-4B}, trained from the base model with the unmasked user for 90 steps; \texttt{gpt-oss-120b} as the user in both stages), against a control trained with the task reward alone on the same data. Neither bonus helped on 100 held-out tasks (\autoref{tab:repair}), and neither changed behaviour: repair moves were as frequent as in the control, the share of rollouts receiving the bonus did not rise during training (A: 37\%, 25\%, 30\%; B: 14\%, 11\%, 15\%), and the training reward was slightly lower (by 1.5 and 1.3 points over steps 1--32). Only 6.6\% (A) and 1.9\% (B) of groups received a signal from the bonus that differed from the task reward, and 36\% of A's bonuses went to routine questions after the opening message rather than to repairs. A reward for repair behaviour is too sparse to teach it.

\begin{table}[h]
\centering
\small
\setlength{\tabcolsep}{3pt}
\begin{tabular}{lcccc}
\toprule
 & \multicolumn{2}{c}{gpt-oss-20b} & \multicolumn{2}{c}{gpt-oss-120b} \\
\cmidrule(lr){2-3}\cmidrule(lr){4-5}
 & weak & strong & weak & strong \\
\midrule
A $-$ control & $+$1.4\sd{2.4} & $+$0.7\sd{2.7} & $-$0.9\sd{2.0} & $+$1.3\sd{2.4} \\
B $-$ control & $+$0.9\sd{2.2} & $+$1.8\sd{2.9} & $-$1.4\sd{2.9} & $+$0.5\sd{2.9} \\
\bottomrule
\end{tabular}
\caption{Repair bonus minus control (\texttt{Qwen3.5-4B}; points; task-paired, $\pm$ half-width of an approximate 95\% interval) on 100 held-out \texttt{ToolWOZ} tasks, by evaluation user; weak: $p{=}0.4$, 8 repetitions (B with \texttt{gpt-oss-120b}: 4); strong: $p{=}0$, 4 (2).}
\label{tab:repair}
\end{table}

\paragraph{Rewarding the state.} We added a state score to the reward, $R=R_{\text{task}}+\lambda R_{\text{state}}$ with $\lambda{=}1/(4n)$ for a dialogue of $n$ turns, where a grader scores the state block the agent writes at each turn (\texttt{Qwen3.5-2B} from the base model, 16 turns, 60 steps; from step 16 a state with missing fields counts as a format failure). Against a \textsc{GRPO} control with the same configuration and task order, the step-paired training reward is 6.6$\pm$1.0 points lower over 60 steps (7.6, 1.4, 6.8 and 10.6 in epochs 1--4), and the validation score at step 60 is 59.1 against 65.5 (earlier protocol, 21 tasks $\times$ 4). Within a group the state score is unrelated to task success (correlation $-$0.07 to $+$0.04; at most 0.11 for any of its components), so adding it flips the sign of the advantage for 13--17\% of rollouts and adds noise to the task signal. The model learns the format (state score 0.43 to 0.54) but not the content: known parameters are written correctly in 66--69\% of turns throughout, and unseen fields listed as missing fall from 7\% to 5\%. The same reward lowers the training reward on \taub{} as well ($-$6.6$\pm$4.6), and from an SFT model that already writes states it is level with the control over 15 steps ($+$0.6$\pm$1.4).

\paragraph{Writing the state once.} Writing the state only once, after the dialogue, with an advantage computed separately from the task reward, teaches the state (valid format in 92--100\% of dialogues; F-score of the final state 0.51 to 0.59) but not the task: over 15 steps the step-paired training reward is 2.8 to 5.0$\pm$1.0 points below the control (1.7 for a null run), and at step 15 the task-paired validation difference is $-$2.4$\pm$1.9, within the difference of a model compared with itself ($+$3.4$\pm$3.5). A state that is rewarded for being correct is either noise for the task reward or irrelevant to it.

\paragraph{From reward to grouping.} These observations point away from supervising the state and towards using it. A written state need not be correct to be useful: rollouts that write the same key are in comparable situations, so their turns can be compared and credited by what happened next (\autoref{sec:method}, \autoref{app:state}). \MethodName therefore uses the state as a grouping key, not as a target.

\FloatBarrier
\section{Further Results for RQ2--3}
\label{app:sg}

\subsection{All \MethodName Runs}
\label{app:sgfull}

\begin{table*}[t]
\centering
\setlength{\tabcolsep}{2pt}
\begin{tabularx}{\textwidth}{l*{6}{>{\centering\arraybackslash}X}}
\toprule
 & \multicolumn{3}{c}{\taub{}} & \multicolumn{3}{c}{\texttt{ToolWOZ}} \\
\cmidrule(lr){2-4}\cmidrule(lr){5-7}
Evaluated at & $p{=}0$ & 0.2 & 0.4 & $p{=}0$ & 0.2 & 0.4 \\
\midrule
Untrained & 41.2\sd{5.3} & 33.0\sd{8.2} & 30.7\sd{4.9} & 49.0\sd{2.4} & 42.6\sd{1.8} & 40.7\sd{3.9} \\
\midrule
\grp{7}{From the SFT cold start} \\
\textsc{GRPO} & 38.0\sd{3.5} & 34.9\sd{10.9} & 39.6\sd{7.0} & 70.5\sd{4.5} & 69.7\sd{5.5} & 62.5\sd{6.6} \\
\textsc{UserRL} & 31.2\sd{5.4} & 27.1\sd{5.4} & 28.1\sd{5.2} & 61.4\sd{0.6} & 60.4\sd{2.4} & 57.9\sd{0.3} \\
\rowcolor{oursbg} \MethodName & \textbf{45.8}\sd{4.3} & \textbf{36.5}\sd{3.0} & \textbf{40.6}\sd{2.2} & \textbf{84.1}\sd{3.2} & \textbf{81.5}\sd{3.8} & \textbf{76.8}\sd{1.7} \\
\midrule
\grp{7}{Strongest baselines, from the base model} \\
\textsc{GRPO} & 34.4\sd{4.0} & 30.2\sd{4.0} & 24.0\sd{7.9} & 78.7\sd{3.2} & 74.3\sd{2.3} & 72.0\sd{4.3} \\
\textsc{InfoPO} & 35.4\sd{7.2} & 37.5\sd{4.8} & 30.2\sd{8.6} & 78.3\sd{6.7} & 75.7\sd{1.0} & 74.0\sd{1.5} \\
\midrule
\grp{7}{Without the cold start} \\
\MethodName & 32.3\sd{7.0} & 31.8\sd{6.7} & 35.4\sd{6.7} & 74.8\sd{0.7} & 72.2\sd{0.9} & 67.1\sd{4.2} \\
\bottomrule
\end{tabularx}
\caption{\autoref{tab:sg} with standard deviations over evaluation rounds (\texttt{Qwen3.5-2B}, trained at $p{=}0.4$).}
\label{tab:sg_full}
\end{table*}

\begin{table}[t]
\centering
\small
\setlength{\tabcolsep}{3.5pt}
\begin{tabular}{lccc}
\toprule
Trained and evaluated at & $p{=}0$ & 0.2 & 0.4 \\
\midrule
\grp{4}{Qwen3.5-2B} \\
\textsc{GRPO} (base) & 74.9\sd{0.2} & \textbf{74.8}\sd{6.3} & 72.0\sd{4.3} \\
SFT, \textsc{GRPO} & 75.8\sd{5.5} & 74.3\sd{3.5} & 62.5\sd{6.6} \\
\textsc{UserRL} (SFT) & 76.5\sd{3.5} & 69.3\sd{0.8} & 57.9\sd{0.3} \\
\textsc{InfoPO} (base) & 85.3\sd{1.3} & 72.4\sd{2.4} & 74.0\sd{1.5} \\
\rowcolor{oursbg} SFT, \MethodName & \textbf{82.3}\sd{7.0} & \textbf{74.8}\sd{1.7} & \textbf{76.8}\sd{1.7} \\
\midrule
\grp{4}{Qwen3.5-4B} \\
\textsc{GRPO} (base) & 85.5\sd{2.9} & 82.4\sd{0.6} & 74.8\sd{2.7} \\
SFT, \textsc{GRPO} & 87.9\sd{1.1} & 87.0\sd{2.8} & 77.2\sd{2.7} \\
\rowcolor{oursbg} SFT, \MethodName & 88.1\sd{1.9} & 87.1\sd{1.7} & 81.5\sd{1.7} \\
\bottomrule
\end{tabular}
\caption{Rows of \autoref{tab:sgmask} with standard deviations over evaluation rounds (\texttt{ToolWOZ}).}
\label{tab:sgmask_full}
\end{table}

\autoref{tab:app_tw} and \autoref{tab:app_rt} list the training runs of \autoref{tab:sg}: the selected checkpoint, its validation score and its test score under every evaluation mask; test scores at every candidate checkpoint on \texttt{ToolWOZ} are in \autoref{fig:bystep_full}.
% On \taur{}, \MethodName rows use the shaped reward and all other rows the binary reward; the base-model \MethodName run began replying without calling tools after step 60 and was stopped, so its candidates end at step 60.

\begin{table}[t]
\centering
\small
\setlength{\tabcolsep}{3.4pt}
\begin{tabular}{llcccc}
\toprule
 & & & \multicolumn{3}{c}{Test at selected step} \\
\cmidrule(lr){4-6}
Method & Train & Val & $p{=}0$ & 0.2 & 0.4 \\
\midrule
\grp{6}{Qwen3.5-2B} \\
\textsc{GRPO} (from base) & 0.4 & 45.0 & 34.4 & 30.2 & 24.0 \\

SFT, \textsc{GRPO} & 0.4 & 57.5 & 38.0 & 34.9 & 39.6 \\

\textsc{InfoPO} (from base) & 0.4 & 45.0 & 35.4 & 37.5 & 30.2 \\

\textsc{UserRL} (SFT) & 0.4 & 35.0 & 31.2 & 27.1 & 28.1 \\

\MethodName (from base) & 0.4 & 52.5 & 32.3 & 31.8 & 35.4 \\

SFT, \MethodName & 0 & 52.5 & 33.3 & 22.9 & 29.2 \\
SFT, \MethodName & 0.2 & 47.5 & 27.1 & 32.3 & 28.1 \\

SFT, \MethodName & 0.4 & 47.5 & 45.8 & 36.5 & 40.6 \\
\midrule
\grp{6}{Qwen3.5-4B} \\
SFT, \textsc{GRPO} & 0.4 & 60.0 & 38.5 & 33.3 & 37.5 \\
SFT, \MethodName & 0 & 52.5 & 45.8 & 43.8 & 39.6 \\
SFT, \MethodName & 0.2 & 45.0 & 44.8 & 50.0 & 44.8 \\
SFT, \MethodName & 0.4 & 57.5 & 51.0 & 44.8 & 42.7 \\
% SFT, \MethodName & 0.4 & 47.5 & -- & -- & 40.6 \\

\bottomrule
\end{tabular}
\caption{\taur{}: training runs, as in \autoref{tab:app_tw}; validation on the 10 development tasks, test on the 24 test tasks $\times$ 4 seeds; \texttt{Qwen3.5-4B} runs compared at steps 20 and 40, except \MethodName at mask 0.4 (steps 20--60) and SFT, \textsc{GRPO} (steps 20--80). }
% $^*$Run or evaluation not finished.}
\label{tab:app_rt}
\end{table}

\begin{table}[t]
\centering
\small
\setlength{\tabcolsep}{3.4pt}
\begin{tabular}{llcccc}
\toprule
 & & & \multicolumn{3}{c}{Test at selected step} \\
\cmidrule(lr){4-6}
Method & Train & Val & $p{=}0$ & 0.2 & 0.4 \\
\midrule
\grp{6}{Qwen3.5-2B} \\
\textsc{GRPO} (from base) & 0.4 & 66.8 & 82.3 & 74.7 & 75.2 \\
\textsc{GRPO} (from base) & 0.4 & 67.5 & 76.2 & 75.7 & 69.1 \\
\textsc{GRPO} (from base) & 0.4 & 68.5 & 77.6 & 72.6 & 71.8 \\
\textsc{InfoPO} (from base) & 0 & 76.9 & 85.3 & 78.3 & 74.8 \\
\textsc{InfoPO} (from base) & 0.2 & 67.4 & 75.4 & 72.4 & 69.0 \\
\textsc{InfoPO} (from base) & 0.4 & 62.1 & 78.3 & 75.7 & 74.0 \\
\MethodName (from base) & 0.4 & 64.7 & 74.8 & 72.2 & 67.1 \\
\addlinespace
SFT, \textsc{GRPO} & 0 & 70.8 & 80.1 & 69.9 & 68.7 \\
SFT, \textsc{GRPO} & 0 & 74.2 & 71.5 & 67.9 & 65.4 \\
SFT, \textsc{GRPO} & 0.2 & 77.8 & 81.5 & 74.3 & 66.0 \\
SFT, \textsc{GRPO} & 0.4 & 66.7 & 74.8 & 75.7 & 69.6 \\
SFT, \textsc{GRPO} & 0.4 & 61.9 & 67.7 & 63.8 & 55.8 \\
SFT, \textsc{GRPO} & 0.4 & 61.8 & 69.0 & 69.5 & 62.0 \\
\textsc{UserRL} (SFT) & 0 & 69.6 & 76.5 & 69.8 & 66.4 \\
\textsc{UserRL} (SFT) & 0.2 & 73.9 & 72.9 & 69.3 & 60.7 \\
\textsc{UserRL} (SFT) & 0.4 & 52.8 & 61.4 & 60.4 & 57.9 \\
SFT, \MethodName & 0 & 81.5 & 88.1 & 78.6 & 73.6 \\
SFT, \MethodName & 0 & 83.2 & 76.4 & 76.2 & 72.2 \\
SFT, \MethodName & 0.2 & 73.1 & 76.9 & 74.8 & 71.6 \\
SFT, \MethodName & 0.4 & 75.7 & 84.3 & 82.0 & 77.4 \\
SFT, \MethodName & 0.4 & 72.8 & 81.5 & 78.6 & 75.7 \\
SFT, \MethodName & 0.4 & 70.1 & 86.6 & 83.8 & 77.4 \\
\midrule
\grp{6}{Qwen3.5-4B} \\
SFT, \textsc{GRPO} & 0 & 87.6 & 87.9 & 86.6 & 80.6 \\
SFT, \textsc{GRPO} & 0.2 & 80.4 & 86.2 & 87.0 & 80.4 \\
SFT, \textsc{GRPO} & 0.4 & 72.1 & 85.6 & 82.4 & 77.2 \\
\textsc{UserRL} (SFT) & 0 & 80.5 & 86.1 & 81.0 & 70.3 \\
\textsc{UserRL} (SFT) & 0.2 & 55.6 & 59.3 & 48.7 & 43.3 \\
\textsc{UserRL} (SFT) & 0.4 & 64.2 & 69.2 & 66.0 & 59.1 \\
\textsc{InfoPO} (from base) & 0 & 84.6 & 87.9 & 85.4 & 77.8 \\
\textsc{InfoPO} (from base) & 0.2 & 71.4 & 82.3 & 82.1 & 78.4 \\
\textsc{InfoPO} (from base) & 0.4 & 71.4 & 83.8 & 84.6 & 77.9 \\
SFT, \MethodName & 0 & 85.7 & 88.1 & 83.9 & 80.8 \\
SFT, \MethodName & 0.2 & 87.8 & 89.6 & 87.1 & 82.9 \\
SFT, \MethodName & 0.4 & 74.9 & 87.1 & 85.2 & 81.5 \\
\bottomrule
\end{tabular}
\caption{ToolWOZ: training runs. Val: validation score (30 tasks, training mask) of the checkpoint selected among steps 30/60/90/120; rows with the same method and mask are separate training runs; test: 70 tasks $\times$ 2 under each evaluation mask.}
\label{tab:app_tw}
\end{table}

\subsection{Training Without the Cold Start}
\label{app:nocold}

The main text trains \MethodName and \textsc{UserRL} from the SFT cold start and \textsc{GRPO} from the base model (\autoref{sec:expsetup}). \autoref{tab:nocold} reports \textsc{GRPO}, \textsc{InfoPO} and \MethodName trained from the base model, \textsc{InfoPO} as in its paper, with the cold-started \MethodName for reference. Without the cold start \textsc{GRPO} is stronger than with it (72.0 vs.\ 62.5 on \texttt{ToolWOZ} at $p{=}0.4$), and \MethodName is weaker (67.1 vs.\ 76.8) and not above \textsc{GRPO}. A plausible explanation is that the cold start narrows exploration, and \textsc{GRPO}, which credits every turn of a dialogue alike, then reinforces repeated tool calls (\autoref{app:failmode}), whereas the step-level credit of \MethodName ranks a turn that loops below one that makes progress from the same situation. Trained and evaluated at masks 0 and 0.2, \textsc{InfoPO} scores 85.3 and 72.4 on \texttt{ToolWOZ}; \texttt{Qwen3.5-4B} \textsc{GRPO} from the base model (the arms of \autoref{tab:main}, learning rate $3{\times}10^{-5}$) scores 85.5, 82.4 and 74.8 at masks 0, 0.2 and 0.4. \autoref{tab:user_nocold} gives the interaction metrics of \autoref{fig:user} for these runs.

\begin{table}[t]
\centering
\small
\setlength{\tabcolsep}{2.2pt}
\begin{tabular}{lcccccc}
\toprule
 & \multicolumn{3}{c}{\taub{}} & \multicolumn{3}{c}{\texttt{ToolWOZ}} \\
\cmidrule(lr){2-4}\cmidrule(lr){5-7}
Eval.\ at & $p{=}0$ & 0.2 & 0.4 & $p{=}0$ & 0.2 & 0.4 \\
\midrule
\textsc{GRPO} & 34.4 & 30.2 & 24.0 & 78.7 & 74.3 & 72.0 \\
\textsc{InfoPO} & 35.4 & 37.5 & 30.2 & 78.3 & 75.7 & 74.0 \\
\MethodName & 32.3 & 31.8 & 35.4 & 74.8 & 72.2 & 67.1 \\
\midrule
\rowcolor{oursbg} \MethodName{} (SFT) & 45.8 & 36.5 & 40.6 & 84.1 & 81.5 & 76.8 \\
\bottomrule
\end{tabular}
\caption{\textbf{Without the cold start} (\texttt{Qwen3.5-2B}, trained from the base model with the weak user, $p{=}0.4$, as in \autoref{tab:sg}); last row: \MethodName with the cold start, from \autoref{tab:sg}. Standard deviations: \autoref{tab:sg_full}.}
\label{tab:nocold}
\end{table}

\begin{table}[t]
\centering
\small
\setlength{\tabcolsep}{3.2pt}
\begin{tabular}{lccccc}
\toprule
 & Succ. & Agent & User & Turn & Ques- \\
 & (\%)$\uparrow$ & turns$\downarrow$ & replies$\downarrow$ & limit$\downarrow$ & tions \\
\midrule
\textsc{GRPO} & 72.0 & 12.3 & 5.8 & 38 & 3.5 \\
\textsc{InfoPO} & 74.0 & \textbf{11.6} & \textbf{4.1} & \textbf{22} & 3.9 \\
\MethodName & 67.1 & 12.9 & 6.3 & 49 & 2.9 \\
\bottomrule
\end{tabular}
\caption{Interaction with the weak user without the cold start (\texttt{ToolWOZ}, \texttt{Qwen3.5-2B}, trained and evaluated at $p{=}0.4$), as in \autoref{fig:user}.}
\label{tab:user_nocold}
\end{table}

\begin{table}[t]
\centering
\small
\setlength{\tabcolsep}{4pt}
\begin{tabular}{lcccc}
\toprule
 & \multicolumn{2}{c}{\texttt{ToolWOZ}} & \multicolumn{2}{c}{\taub{}} \\
\cmidrule(lr){2-3}\cmidrule(lr){4-5}
Cold start & none & SFT & none & SFT \\
\midrule
\textsc{GRPO} & \textbf{72.0} & 62.5 & 24.0 & 39.6 \\
\rowcolor{oursbg} \MethodName & 67.1 & \textbf{76.8} & \textbf{35.4} & \textbf{40.6} \\
\midrule
Anchored, first 20 steps & 24\% & 30\% & 28\% & 35\% \\
Anchored, last 20 steps & 19\% & 48\% & 20\% & 48\% \\
\bottomrule
\end{tabular}
\caption{\textbf{Effect of the cold start} (\texttt{Qwen3.5-2B}, trained and evaluated at $p{=}0.4$). Anchored turns: share of \MethodName training turns that fall into an anchor group (SFT start: the higher of the two runs with complete logs).}
\label{tab:cold}
\end{table}

\paragraph{Effect of the cold start.} Without the cold start, \MethodName is below \textsc{GRPO} on \texttt{ToolWOZ} (67.1 vs.\ 72.0, within two standard errors) and above it on \taub{} (35.4 vs.\ 24.0). With it, \MethodName is above \textsc{GRPO} on both benchmarks (\autoref{tab:cold}). The anchors depend on the states the agent writes. On \texttt{ToolWOZ}, from the SFT start the share of turns that fall into an anchor group rises from 30\% in the first 20 steps to 48\% in the last 20, with 21--30 anchor groups per step; from the base model it falls from 24\% to 19\%, with 8.5 groups per step. On \taub{} it rises from 35\% to 48\% from the SFT start and falls from 28\% to 20\% from the base model. The cold start makes the states written by different rollouts coincide, so about twice as many turns receive a step-level signal. We report this as an observation; we did not vary the cold start in other ways.

\subsection{Grouping versus Writing the State}
\label{app:ablation}

\autoref{tab:ablation} separates the two parts of \MethodName. All three configurations start from the same SFT cold start and use the same training settings and checkpoint selection; they differ only in whether the agent writes the state and whether turns are grouped by it. With the state but without grouping ($\omega{=}0$), \textsc{GRPO} is no better than without the state, and \MethodName is 15--17 points above it and 12--14 points above \textsc{GRPO} without the state under every evaluation mask. Without grouping, training is also less stable: one of the three runs collapsed after step 80 (KL to the reference policy 0.54, test success about 10 at step 90), and its selected checkpoint is step 30; none of the three \MethodName runs collapsed.

\begin{table}[h]
\centering
\small
\setlength{\tabcolsep}{3pt}
\begin{tabular}{lcccc}
\toprule
State & Grouping & $p{=}0$ & 0.2 & 0.4 \\
\midrule
written & no ($\omega{=}0$) & 69.0\sd{5.6} & 66.6\sd{2.1} & 60.1\sd{3.6} \\
\rowcolor{oursbg} written & yes (\MethodName) & \textbf{84.1}\sd{2.5} & \textbf{81.5}\sd{2.6} & \textbf{76.8}\sd{1.0} \\
none & no (\textsc{GRPO}) & 70.5\sd{3.8} & 69.7\sd{6.0} & 62.5\sd{6.9} \\
\bottomrule
\end{tabular}
\caption{\textbf{Grouping versus writing the state} (\texttt{ToolWOZ}, \texttt{Qwen3.5-2B}, SFT cold start, trained at $p{=}0.4$, 70 test tasks). Mean $\pm$ standard deviation over three training runs.}
\label{tab:ablation}
\end{table}

\subsection{Training Dynamics}
\label{app:sgfigs}

\begin{figure*}[t]
\centering
\includegraphics[width=0.95\textwidth]{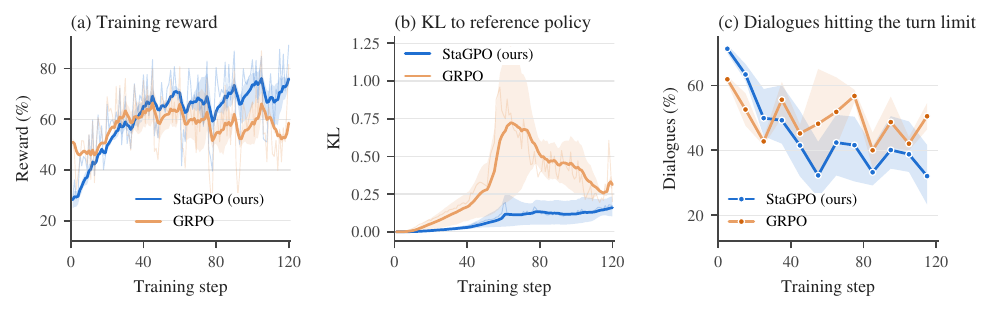}
\caption{\textbf{Training curves of \MethodName and \textsc{GRPO}} (\texttt{ToolWOZ}, \texttt{Qwen3.5-2B}, both from the SFT cold start, trained at $p{=}0.4$). (a) Training reward and (b) KL divergence to the reference policy per training step (thin: mean over runs, bold: smoothed); (c) share of training dialogues that use all 16 agent turns, in bins of 10 steps. Shading: range over runs.}
\label{fig:why}
\end{figure*}

\begin{figure}[t]
\centering
\includegraphics[width=\columnwidth]{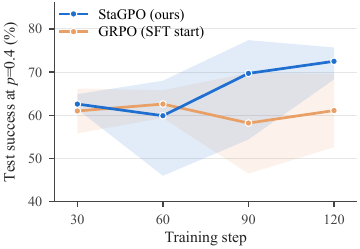}
\caption{Test success at $p{=}0.4$ at every candidate checkpoint (\texttt{ToolWOZ}, \texttt{Qwen3.5-2B}, both from the SFT cold start; band: range over three runs).}
\label{fig:bystep_full}
\end{figure}

\autoref{fig:bystep_full} gives test success under the weak user at every candidate checkpoint on \texttt{ToolWOZ} (mean and range over three training runs); training reward, KL and dialogues that reach the 16-turn limit are in \autoref{fig:why}. In \textsc{GRPO} rollouts about 70\% of dialogues contain an exactly repeated tool call, and these hit the 16-turn limit in 77--84\% of cases; \MethodName does not reduce verbatim repeats, but fewer of its dialogues run into the turn limit (\autoref{fig:why}c). Its KL to the reference policy stays at roughly a quarter of \textsc{GRPO}'s while its training reward keeps rising, and its runs vary less (\autoref{fig:why}a,b).

% Appendix turn-limit figure (now panel (c) of fig:why)
\iffalse
\begin{figure}[h]
\centering
\includegraphics[width=\columnwidth,trim=0 0 251 0,clip]{figures/figC_mechanism.pdf}
\caption{ToolWOZ training rollouts (SFT start): dialogues with a verbatim repeated tool call and dialogues that hit the 16-turn limit.}

\end{figure}
\fi

\subsection{A Failure Mode of \textsc{GRPO}}
\label{app:failmode}

In rollouts of plain \textsc{GRPO} trained at $p{=}0.4$, about 70\% of dialogues contain an exactly repeated tool call. These dialogues hit the 16-turn limit in 77--84\% of cases, against 27--41\% for dialogues without a repeat, and score 35--38 against 53--61. Nothing in the context records that a call failed, so a low-entropy policy repeats it.

\FloatBarrier
\section{Extended Related Work}
\label{app:related}

Large language models (LLMs) have advanced rapidly in recent years \citep{ye2025llms4all, zhang2025agentrouter, xiong2026adaptive, xiong2026dive}. Building on this progress, LLM-driven agents and agentic reinforcement learning have gained increasing attention for their ability to plan, interact with external environments, and solve complex tasks with limited human oversight \citep{shi2026ng, huang2026evolverouter}.

\paragraph{Multi-turn agent RL with simulated users.} Reinforcement learning for multi-turn language agents has moved from hierarchical value methods \citep{zhou2024archer} to trajectory-level policy optimization with verifiable outcome rewards \citep{wang2025ragen}, and to environments in which the other party is an LLM-simulated user: SWEET-RL trains collaborative agents against a simulated human \citep{zhou2025sweetrl}, and \textsc{UserRL} trains tool agents across a suite of user-centric gym environments \citep{qian2025userrl}. Step-level credit without a critic has been obtained by grouping steps that reach an identical environment state across rollouts \citep{feng2025gigpo}, and response-level uncertainty has been used to modulate \textsc{GRPO} updates, where geometry-aware, reward-calibrated estimates outperform semantic entropy \citep{zhang2026semantic}; \MethodName instead modulates credit by the situation a turn is in. These works vary the algorithm, the reward or the task. We hold all three fixed and vary only how much of the user's reply reaches the agent, which isolates the training partner as a variable.

\paragraph{How strong are simulated users?} Tool-agent benchmarks pair the agent with a simulated user that holds the full task: $\tau$-bench and $\tau^2$-bench in customer service \citep{yao2024taubench,barres2025tau2}, \texttt{ToolWOZ} in \texttt{MultiWOZ} booking domains \citep{lattimer2024sparse,budzianowski2018multiwoz}, MINT with simulated language feedback \citep{wang2023mint}, and ToolSandbox with a built-in user simulator and an explicit \emph{insufficient information} category \citep{lu2024toolsandbox}. Dynamic evaluation protocols generate test cases on demand instead of fixing a benchmark \citep{bao2025autodavis,huang2026trustgen}; our masking likewise varies the user while holding the task fixed. Several studies make the user harder. Non-collaborative user simulators add tangents, impatience, unavailable requests and incomplete utterances while still delivering every needed fact, and degrade agents on \texttt{MultiWOZ} and $\tau$-bench \citep{shim2025noncollab}. Sharding a fully specified instruction across turns lowers performance by 39\% on average, mostly because models commit to early assumptions \citep{laban2025lost}. These works change the user's behaviour through prompting and evaluate fixed models. We keep the simulator unchanged and degrade the channel after it speaks, which preserves the task, records what was said and what was heard, and lets us ask whether \emph{training} with the weaker user helps.

\paragraph{Underspecification and clarification.} Asking clarifying questions has a long history in information seeking \citep{aliannejadi2019asking, huang2026cleat}. For language agents, IN3 studies vague task instructions \citep{qian2024tellmemore}; NoisyToolBench finds that tool-calling models fabricate missing arguments and proposes prompting them to ask when needed \citep{wang2024learningtoask}; Ambig-SWE finds that agents struggle to tell underspecified from fully specified tasks, although interaction helps once they do ask \citep{vijayvargiya2025ambigswe}; and Drift-Bench benchmarks show that agents under input faults tend to act rather than ask \citep{bao2026driftbench, zhang2026oracle}. Training methods reward useful questions through self-improvement \citep{andukuri2024stargate}, contrastive self-training on conversational actions \citep{chen2024learning}, simulated future turns \citep{zhang2024modeling}, multi-turn-aware rewards \citep{wu2025collabllm} or turn-level information gain \citep{kong2026infopo}. In these settings the missing information is absent from the request and a well-chosen question retrieves it. In ours the user has said it but it did not arrive, and the trained agent already asks for most lost identifiers; we find that inducing more questions recovers little and that the open problem is recognising which requirement is still unresolved before acting.

\paragraph{Training under perturbation.} Robustness to perturbed inputs is one of the dimensions on which the trustworthiness of generative models is benchmarked, with test cases varied dynamically \citep{huang2026trustgen}. Robustness to noisy user input has been studied for language understanding in task-oriented dialogue, including speech-like errors and noise perturbations \citep{liu2020laug}. The standard remedy for a train--test mismatch is to train under the perturbation, as in domain randomisation \citep{tobin2017domain}; for LLM agents, NoisyAgent injects environmental imperfections including user ambiguity, combining mixed and curriculum schedules \citep{chen2026noisyagent}. These studies emphasize identifying uncertainty introduced by imperfect observations or execution conditions so that agents can trigger effective and efficient interventions \citep{zhang2026optimizing}. We test fixed, mixed and curriculum schedules under a controlled user-side degradation. Unlike perceptual noise, our perturbation removes information the reward depends on, and none of these schedules serves weak users better than training with strong users.

\paragraph{Privileged information.} Information available only during training can shape a learner that never sees it at test time \citep{vapnik2009lupi}; asymmetric actor-critic methods give the critic the full state \citep{pinto2017asymmetric}, and SWEET-RL gives a turn-level critic access to the reference solution \citep{zhou2025sweetrl}. \MethodName (\autoref{sec:method}) instead compares turns that reach the same state written by the agent itself, a step-level baseline that needs no learned critic.

\FloatBarrier
\end{document}